%% file: main.tex
\documentclass[10pt,twocolumn,letterpaper]{article}

\usepackage[pagenumbers]{cvpr} 
\usepackage{booktabs} 
\usepackage{siunitx}  
\usepackage{multirow} 
\usepackage{array}    
\usepackage{graphicx}
\usepackage{lineno}
\usepackage[table]{xcolor}
\usepackage{listings}
\input{preamble}

\definecolor{cvprblue}{rgb}{0.21,0.49,0.74}
\usepackage[pagebackref,breaklinks,colorlinks,allcolors=cvprblue]{hyperref}

\def\paperID{} 
\def\confName{CVPR}
\def\confYear{2026}

\title{Beyond Static Artifacts: A Forensic Benchmark for Video \\ Deepfake Reasoning in Vision Language Models}

\author{
    Zheyuan Gu$^{1,2*}$\quad 
    Qingsong Zhao$^{1,3*}$\quad 
    Yusong Wang$^{1*}$\quad 
    Zhaohong Huang$^1$ \\
    Xinqi Li$^{2}$\quad 
    Cheng Yuan$^1$\quad 
    Jiaowei Shao$^1$\quad 
    Chi Zhang$^1$\quad 
    Xuelong Li$^{1\dagger}$ \\
    \and
    $^1$Institute of Artificial Intelligence, China Telecom (TeleAI)\quad $^2$Peking University\quad
    $^3$Fudan University\\
}

\begin{document}
\maketitle
\def\thefootnote{*}\footnotetext{Equal contribution.}
\def\thefootnote{$\dagger$}\footnotetext{Corresponding author.}
\def\thefootnote{\arabic{footnote}} 

\input{sec/abstract}

\input{sec/introduction}

\input{sec/related_work}

\input{sec/methodology}

\input{sec/experiment}

\input{sec/conclusion}

{
    \small
    \bibliographystyle{ieeenat_fullname}
    \bibliography{main}
}

\input{sec/X_suppl}

\end{document}

%% file: preamble.tex
\newcommand{\Bench}{FAQ}
\newcommand{\BenchIT}{FAQ-IT}
\newcommand{\LI}{\textit{Facial Perception}}
\newcommand{\LII}{\textit{Temporal Deepfake Grounding}}
\newcommand{\LIII}{\textit{Forensic Reasoning}}
\usepackage{subcaption}
\usepackage{graphicx}
\newcommand{\good}[1]{\textbf{\textcolor{AcademicGreen}{#1}}} 
\newcommand{\bad}[1]{\textbf{\textcolor{red}{#1}}}
\definecolor{AcademicGreen}{RGB}{0, 100, 0}  

%% file: sec/abstract.tex
\begin{abstract}
Current Vision-Language Models (VLMs) for deepfake detection excel at identifying spatial artifacts but overlook a critical dimension: temporal inconsistencies in video forgeries.
Adapting VLMs to reason about these dynamic cues remains a distinct challenge. 
To bridge this gap, we propose Forensic Answer-Questioning (FAQ), a large-scale benchmark that formulates temporal deepfake analysis as a multiple-choice task.
FAQ introduces a three-level hierarchy to progressively evaluate and equip VLMs with forensic capabilities: 
(1) Facial Perception, testing the ability to identify static visual artifacts;
(2) Temporal Deepfake Grounding, requiring the localization of dynamic forgery artifacts across frames;
and (3) Forensic Reasoning, challenging models to synthesize evidence for final authenticity verdicts.
We evaluate a range of VLMs on FAQ and generate a corresponding instruction-tuning set, FAQ-IT.
Extensive experiments show that models fine-tuned on FAQ-IT achieve advanced performance on both in-domain and cross-dataset detection benchmarks.
Ablation studies further validate the impact of our key design choices, confirming that FAQ is the driving force behind the temporal reasoning capabilities of these VLMs.
\end{abstract}
\vspace{-0.6cm}

%% file: sec/introduction.tex
\vspace{3mm}

\section{Introduction}
\label{sec:intro}

With the rapid rise of AIGC \citep{gan,rombach2022highresolutionimagesynthesislatent,kang2023scalingganstexttoimagesynthesis,scaleupgan,marra2018gansleaveartificialfingerprints}, the creation of realistic deepfakes has become significantly easier, leading to growing social concerns about their potential risks \citep{Deepfake_in_finance,Deepfake_in_politics}.
Consequently, researchers have begun exploring Vision–Language Models (VLMs) for deepfake detection, leveraging their strong visual understanding acquired through large-scale pre-training \citep{qwen25vl,gpt4v} and achieved a number of encouraging results.
FakeRadar and DeepShield \citep{fakeradar, deepshield} adapt CLIP \citep{CLIP} for deepfake detection, with the former using contrastive learning and the latter employing anomaly detection to enhance generalization.
To achieve better interpretability, some researchers have also employed larger VLM models for deepfake detection.
SIDA \citep{sida} and FakeShield \citep{fakeshield} constructed the SID-Set and MMTD-Set, respectively.
Through training on these datasets, VLMs can localize and describe forgeries while demonstrating remarkable detection performance.
These early successes demonstrate the potential of using VLMs for deepfake detection, suggesting that larger, more challenging training data is expected to further enhance this capability.

\begin{figure*}[t]
    \centering 
    \includegraphics[width=\textwidth]{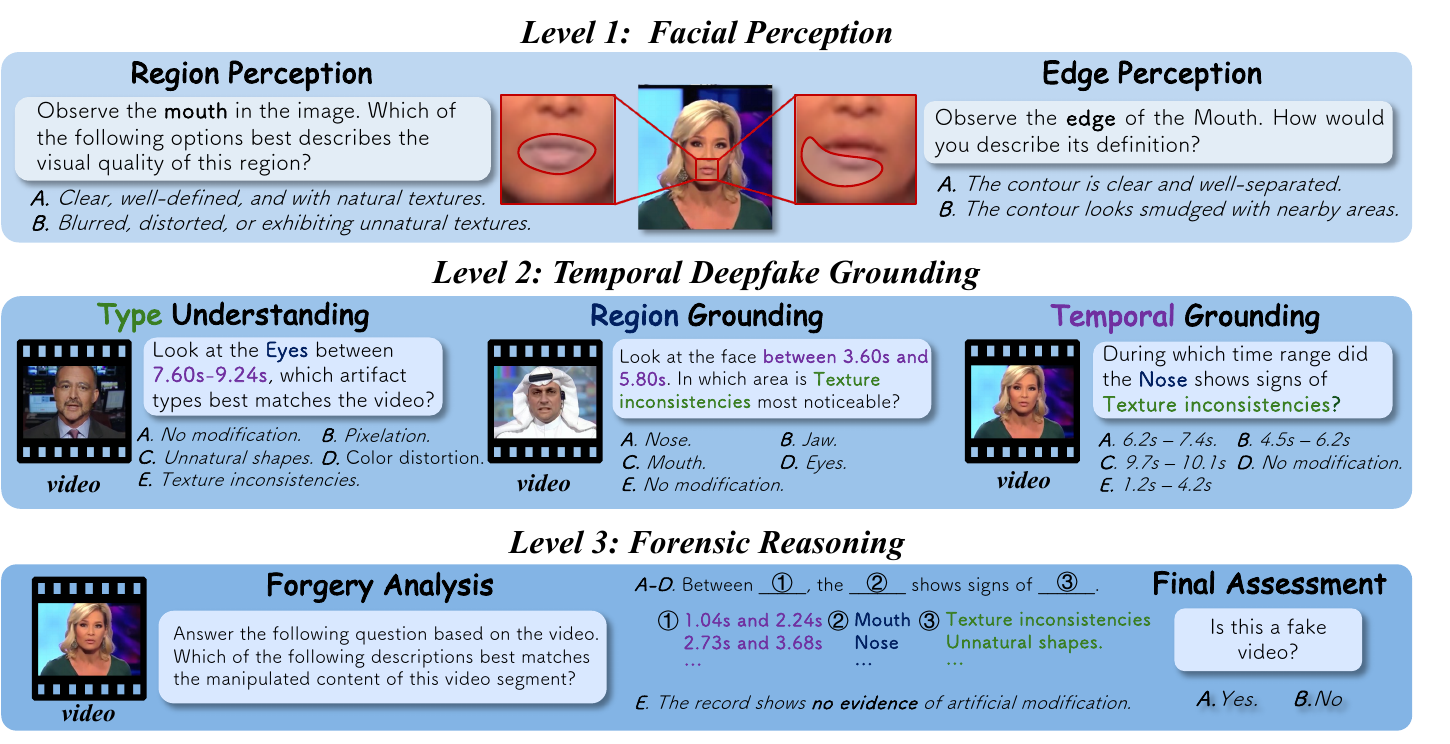} 
    \caption{Illustration of our hierarchical benchmark.}
    \label{fig:benchmark_overview} 
    \vspace{-0.5cm}
\end{figure*}

We therefore ask: \textit{Which VLM training paradigm is most effective for deepfake detection tasks? What constitutes the most promising training data?}
The mainstream approach is supervised fine-tuning (SFT) \citep{sft} with high-quality question answering (QA) datasets.
Previous works \citep{vlffd, ddvqa} extract images from videos and generate annotations using limited question templates and predefined forged regions.
We observed that data obtained from such annotations provides spatial discriminative information only, whereas temporal inconsistencies in deepfake videos have been widely adopted as an important signal in conventional deepfake detection methods \citep{altfreezing, fakesformer, dfgdcg, styleflow}.
We therefore speculate that VLMs trained solely on such data are unable to leverage the significant temporal cues.
One remaining issue is \textit{how to effectively guide VLMs to discover and reason about temporal inconsistencies via constructing QA training data.}

In this work, we continue the vein of SFT-based learning, aiming to establish a comprehensive and robust baseline for the deepfake detection task.
We introduce \underline{F}orensic \underline{A}nswer-\underline{Q}uestioning (FAQ), a multiple-choice question (MCQ) benchmark designed to improve the performance of VLMs in deepfake detection.
We extract clips from videos based on human annotations, build QA pairs through a reproducible automated process, and include carefully designed distractors in the available options.
We construct hierarchical data, which gradually enhances the model's reasoning capabilities for deepfake detection. 
Comparison with other deepfake benchmark in \autoref{tab:benchmark_compare}.

Specifically, questions in FAQ are formulated in three levels to guide VLMs for accurate deepfake detection. Facial perception (level 1) strengthens the model's perception of fine-grained visual artifacts. Temporal deepfake grounding (level 2) improves its grounding ability for dynamic artifacts. Forensic reasoning (level 3) elevates its reasoning capabilities to reach an accurate conclusion.
FAQ contains 33K QA pairs across approximately 4,500 manipulated videos.
Our main contributions are:
\begin{itemize}
    \item To our best knowledge, FAQ is the first QA benchmark focused on temporal inconsistencies in deepfake videos.
    \item We construct a comprehensive QA generation pipeline that leverages static human annotations to locate video segments exhibiting dynamic artifacts and then transform them into QA pairs.
    \item Through extensive experiments, we demonstrate that converting temporal inconsistencies into QA pairs is a feasible VLM training paradigm for deepfake detection. 
Models trained on FAQ show performance gains on widely used detection benchmarks.
\end{itemize}

\begin{table}[t]
\centering
\caption{Comparison between FAQ and other forgery detection benchmarks. H and A respectively indicate human and automatic annotation.}
\label{tab:benchmark_compare}
\footnotesize
\setlength\tabcolsep{5pt}
\resizebox{1.0\columnwidth}{!}{%
\begin{tabular}{llllll}
\toprule
Benchmark & MCQ & \#Sample & Annotation & \#Task Types & Trace \\
\midrule
\multicolumn{6}{@{}c}{\textit{Image-based Benchmarks}} \\
\midrule
GenImage \citep{genimage}  & - & 2,681,167 & A & 1 & - \\
DDVQA \citep{ddvqa} & - & 14,784 & H & 1 & - \\
VLFFD \citep{vlffd} & - & 25,663 & A+H & 1 & - \\
MMTD-Set \citep{fakeshield} & - & 34,470 & A & 2 & - \\
SID-Set$_{desc}$ \citep{sida} & - & 3,000 & A & 2 & - \\
\midrule
\multicolumn{6}{@{}c}{\textit{Video-based Benchmarks}} \\
\midrule
GenVideo \citep{genvideo} & - & 2,294,594 & A & 1 & - \\
Forensics-Bench \citep{Forensics-Bench} & \checkmark & 63,292 & A+H & 4 & - \\
\midrule
FAQ & \checkmark & 33,000 & A+H & 7 & \checkmark \\
\bottomrule
\end{tabular}}
\vspace{-10mm}
\end{table}

%% file: sec/related_work.tex
\section{Related Work}
\label{sec:relatedwork}

\subsection{Traditional Deepfake Detection Datasets}
Early research on deepfake detection datasets focused on binary classification of manipulated and authentic videos.
The most widely adopted benchmark is FaceForensics++ (FF++) \citep{FF++}, which contains videos tampered by face-swapping and reenactment methods under different compression levels.
To enhance realism and data diversity, Celeb-DF \citep{cdf} employed an improved synthesis pipeline to generate visually convincing manipulations common in online media.
Similarly, DeeperForensics \citep{dfo} emphasized data diversity by applying extensive perturbations such as compression, blurring, and color distortion to simulate real-world degradations.
Beyond curated studio-quality samples, WildDeepfake \citep{wild_deepfake} directly collected videos from the Internet with uncontrolled variations in pose, lighting, and quality, challenging the generalization capability of detection methods.
While these traditional datasets are instrumental for developing detection methods, they do not explicitly explain why or where a video is fake.
As a result, models trained on such data often lack interpretability and reasoning capability, especially in cross-domain scenarios.


\subsection{VLMs for Multimedia Forensics}
Recent advances in VLMs have inspired a new paradigm for multimedia forensics, where deepfake detection is no longer treated as a simple classification problem but as a multimodal reasoning task combining visual perception and textual interpretation \citep{fakeshield,sida,M2F2_det}.
FakeShield \citep{fakeshield} introduced an explainable framework for image forgery detection and localization, which leverages VLMs to provide interpretable textual reasoning and pixel-level localization rather than binary predictions.
Huang et al. \citep{sida} extended deepfake image detection to the social media scenario and proposed a unified model that detects, localizes, and explains potential manipulations in online content.
Concurrently, M2F2-Det \citep{M2F2_det} examined the role of contrastive language-image pretraining in face forgery detection.
This model conducts prompt-driven language reasoning based on CLIP representations to produce both authenticity judgments and detailed explanations.
However, these methods \citep{fakeshield,sida,M2F2_det} treat videos as a collection of static frames and largely neglect the temporal evidence on manipulations.
In contrast, our method focuses on dynamic inconsistencies in the time domain and constructs temporal grounding and reasoning QA pairs to equip VLMs with the capabilities to discover this evidence.

\vspace{-1mm}

\subsection{QA-based Deepfake Detection Datasets}

The adoption of VLMs in deepfake detection has stimulated the development of QA datasets.
DD-VQA \citep{ddvqa} is the first QA-based face forgery dataset on human perceptual anomalies such as lighting, texture, and facial inconsistency, with the objective of producing interpretable analysis.
VLFFD \citep{vlffd} further employs an automatic prompt generation pipeline to synthesize large-scale QA pairs of higher quality, enabling coarse-to-fine supervision for face forgery detection.
Forensics-Bench \citep{Forensics-Bench} includes diverse tasks such as recognition, localization, and reasoning, with a focus on synthetic videos from generative models.
Existing datasets are restricted to static images \citep{ddvqa,vlffd} or AI-generated content \citep{Forensics-Bench}.
Conversely, our dataset guides VLMs to detect deepfake videos through curated three-layer QA pairs.

%% file: sec/methodology.tex
\section{FAQ Benchmark }
\label{sec:method}
\begin{figure*}[t]
    \centering 
    \includegraphics[width=1.0\textwidth]{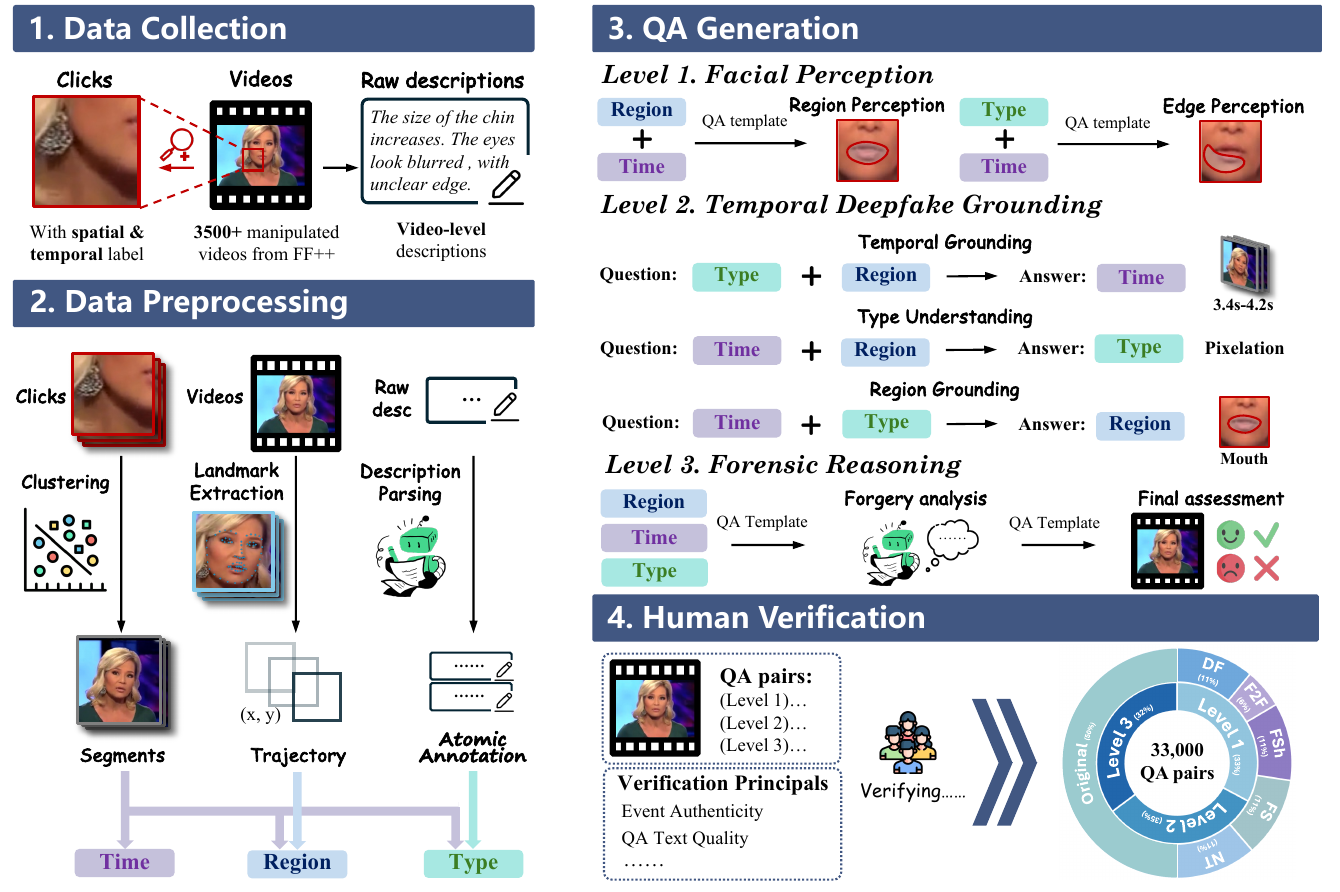} 
    \caption{An overview of \Bench's data construction pipeline.}
    \label{fig:pipeline}
    \vspace{-0.3cm}
\end{figure*}

We develop a benchmark to assess the performance of video-language models on the critical task of detecting video forgeries. 
Specifically, we focus on the temporal inconsistencies inherent in manipulated footage, a non-trivial yet underexplored dimension. 
FAQ is a large-scale MCQ benchmark that evaluates and enhances the capabilities of VLMs in perceiving, grounding, and reasoning about temporal artifacts.
An overview of our benchmark construction pipeline is shown in \autoref{fig:pipeline}.

\subsection{Deepfake Video Curation}
\label{sec:Curation}
\noindent \textbf{Collection.}
We collect 5,000 deepfake videos and 1,000 authentic videos from FaceForensics++ (i.e., FF++ \cite{FF++}), a widely-used benchmark for facial forgery detection.
FF++ provides diverse forgeries with high visual quality, yielding challenging and well-structured video samples.

For each fake video, we augment the dataset with manual annotations, including video-level descriptions and precise spatiotemporal clicks $\mathcal{C}=\{c_i | c_i :=(x_{i},y_{i},t_{i})\}$ that pinpoint forgery artifacts, where $c_i$ is the $i$-th click, $(x_i, y_i)$ are its spatial coordinates, and $t_i$ is the timestamp.
The fine-grained annotations allow us to design challenging forensic reasoning tasks that bridge low-level authenticity discrimination to high-level deepfake analysis.

\noindent \textbf{Filtering.}
For quality control, we deploy YOLOv8 \citep{yolov8} to detect faces in every frame of each fake video and filter out low-quality, easily identifiable samples. 
Specifically, we calculate the average, minimum, and maximum confidence scores of face detection across all frames for each video.

A video is retained only if its average face confidence score exceeds 0.78 and its minimum confidence is above 0.71, ensuring consistent and high-quality facial presence throughout.
This rigorous filtering process removes approximately 10\% of unsatisfactory samples, yielding a refined set of over 4500 high-quality fake videos for subsequent use. 
Detailed statistics, including the sample distribution across confidence intervals, are reported in the Appendix.

\subsection{Preprocessing}
\label{sec:Preprocessing}
For a systematic evaluation of video authenticity,
we structure the raw annotations into three hierarchical tags, serving as the foundational ground truth for generating FAQ. 

\noindent \textbf{Spatio-Temporal Clustering.}
We obtained a total of 14,392 remarkable forged video segments (average duration 2.1 seconds) utilizing over 50,000 sparse clicks provided by human annotators.
Specifically,
we aggregate the sparse clicks into manipulated segments that contain salient dynamic artifacts by grouping those that are spatiotemporally proximate.
This intuitive clustering approach converts discrete annotations into coherent forgery segments.
In view of this, 
we define a spatio-temporal adjacency function $f(c_i, c_j)$ that evaluates to true when two clicks $c_i$ and $c_j$ are within a spatial threshold $\tau_s$ and a temporal threshold $\tau_t$, 
indicating that they belong to the same video segment.
The function is formally expressed as:
\begin{equation}
    f(c_{i},c_{j})=\left(\left\| c_{i}-c_{j}\right\|_2\leq\tau_{s}\right)\land\left(\left| | c_{i}-c_{j} |\right |_1\leq\tau_{t}\right) ,
    \label{eq:eq2}
\end{equation}
where $\left\| \cdot \right\|_2$ and $\left\| \cdot \right\|_1$ correspond to the Euclidean and Manhattan distances, respectively.
In practice, we set $\tau_s = 4$ and $\tau_t = 1$.
Following clustering, one group of clicks defines an individual segment.
The segment's temporal boundaries $(t_i, t_j)$ are determined directly from the earliest and latest timestamps within its click set as the start and end points of a video segment.
For any click that cannot be grouped,
we generate a segment with a symmetric temporal padding of $\delta$ seconds (e.g., $\delta = 0.5s$), resulting in the window $(t_i - \delta, t_i + \delta)$, to capture the complete dynamic artifact.

\noindent \textbf{Landmark Extraction.}
From the 14,392 forged video clips obtained in the previous section, 
we obtain 71,960 trajectories of the manipulated regions.
We extract frame facial landmarks from a video clip using dlib \cite{dlib} and track them to obtain the motion trajectory of a specific facial component.
The facial landmarks (i.e., $\mathcal{P} \in \{p_1,p_2,\cdots,p_{5}\}$) consist of five dominant facial components: eyes, nose, mouth, jaw, and ears.
For each video clip spanning the interval $(t_{i}, t_{j})$, we compute its spatial centroid $\overline{c}_{(i,j)}$ by averaging the coordinates of its constituent clicks.
We then formulate a spatial relevance function that evaluates facial regions based on their distance to the centroid $\overline{c}_{(i,j)}$.

The function computes the total Euclidean distance $S_n$ between the spatial centroid $\overline{c}_{(i,j)}$ and all $\mathcal{P}^{(k)}$ facial landmarks of the $k$-th frame in a video clip, formulated as:
\begin{equation}
	\begin{aligned}
	S_n & = \sum_{k=1}^{K} d\left( \mathcal{P}_n^{(k)},\; \overline{c}_{(i,j)} \right), \\
    n^* & = \arg\min_{n} S_n, \\
    \text{s.t.} & \quad n \in \{1, 2, \cdots, 5\}, \\
	\end{aligned}
    \label{eq:segment_distance}
\end{equation}
where $n$ indexes all facial components, and $n^*$ denotes the forgery-relevant part most associated with the current video clip.
Let $\mathcal{P}^{(k)}_n$  denote the screen coordinate vector of the $n$-th facial component in the $k$-th frame.
The geometric centroids $\mathcal{P}_{n^*}$ from all frames are concatenated to form the forgery trajectory for the current video clip.

\noindent \textbf{Description Parsing. }
We leverage 4000 raw descriptions of the uncut videos to generate 34,921 atomic annotations that specify forgery types,
and subsequently align these fine-grained descriptions with the 14,392 video clips obtained previously.
We begin by performing word frequency analysis (i.e., nltk \cite{nltk}) on raw descriptions to identify common artifact patterns,
leading to the definition of distinct forgery categories such as blurred edges, color distortion, unnatural facial contours, texture inconsistencies, pixelation, and lighting anomalies.
We then design structured prompts to guide an open-source large language model (e.g., gpt-oss-120b \cite{openai2025gptoss120bgptoss20bmodel}) in decomposing the original descriptions into fine-grained, atomic annotations, each describing only a single facial region and one type of artifact.
Finally, we associate these atomic descriptions with corresponding video clips by matching the described facial components to the video clips obtained by clustering above.

\subsection{Task Hierarchy in FAQ}
\label{sec:Task}
Based on the collected forged videos with varying manipulation traces,
we define FAQ's three-level task hierarchy,
benchmarking VLMs on facial forgery videos ranging from fine-grained artifact perception to complex forensic reasoning.
This design enables us to progressively evaluate models' capabilities: whether they can perceive subtle visual inconsistencies,
localize dynamic forgery artifacts across frames,
and ultimately reason about the authenticity of the video.
\autoref{fig:benchmark_overview} illustrates this hierarchical structure.

\noindent \textbf{Level 1: Facial Perception.} This level is designed to evaluate the model's basic perception of visual information in the video through targeted MCQ tasks:
\begin{itemize}
    \item \textbf{Region Perception.}
    Determine whether specific facial regions in the video exhibit clear or blurry visual quality,
    assessing the model's discernment regarding coarse-grained artifacts.
    \item \textbf{Edge Perception.}
    Evaluate whether the model can distinguish between sharp and blurred facial boundaries,
    testing its sensitivity to fine-grained artifacts.
\end{itemize}

\noindent \textbf{Level 2: Temporal Deepfake Grounding.}  
These tasks aim to evaluate the model's ability to localize dynamic artifact cues at spatial and temporal levels, including:
\begin{itemize}
    \item \textbf{Type Understanding.}
    Given a specific time window and facial region,
    determine which type of forgery artifact appears in that region.
    \item \textbf{Region Grounding.}
    Given a temporal segment and artifact type,
    identify the facial region exhibiting such manipulation traces.
    \item \textbf{Temporal Grounding.}
    Provided with potential forgery regions and artifact types,
    localize the time period most likely to contain these artifacts.
\end{itemize}

\noindent \textbf{Level 3: Forensic Reasoning.}  
This level evaluates models' integrated reasoning capacity for holistic deepfake detection through two challenging tasks that require comprehensive analysis of spatiotemporal inconsistencies and evidence synthesis:
\begin{itemize}
    \item \textbf{Forgery Analysis.}
    Requires the model to sequentially identify forgery artifact types,
    locate corresponding facial regions,
    and determine their temporal segments,
    and then select the option that best matches the video content from carefully designed distractors.
    \item \textbf{Final Assessment.}
    After thorough analysis, 
    demand the model to synthesize all forensic evidence and determine the video's authenticity with a conclusive judgment.
\end{itemize}


\subsection{QA Generation with Forged Trace}
\label{annotation}
Our annotation pipeline employs a semi-automated strategy that combines LLM assistance with human instructor supervision.
To maintain focus on dynamic traces, we construct all QA pairs based on the spatiotemporal forgery trajectories derived above, rather than relying on static single-frame descriptions.
In addition,
candidates are carefully designed to be visually and temporally plausible,
forcing models to rely on dynamic visual semantics rather than linguistic cues or LLM priors. To reduce LLM deviation,
the LLM is restricted to generating content only from human-specified dimensions, 
including facial regions, temporal segments, and artifact types.
Human instructors then review, refine, and validate all annotations for correctness.
Building upon the hierarchy introduced in \autoref{sec:Task}, we construct the benchmark as follows.

\noindent \textbf{Facial Perception.}
For the two perception tasks \textit{Region Quality} and \textit{Edge Perception},
candidate captions are first extracted from atomic annotations by LLM.
The human guidance defines the criteria for attribute preservation,
instructing the LLM to retain objective artifact attributes (e.g., blurriness, clarity, key facial areas) while filtering out vague or subjective descriptions.
Then, these resulting captions are converted into four-option MCQs,
where distractors are generated by the LLM following task-specific prompt rules.
The \textit{Region Quality} task is designed to reflect plausible visual alternatives in adjacent facial regions.
Conversely, \textit{Edge Perception} represents common boundary artifacts with similar visual characteristics, forcing models to ground their answers in fine-grained visual signals. 

\noindent \textbf{Temporal Deepfake Grounding.}
For the three grounding tasks of \textit{Type Understanding}, \textit{Region Grounding}, and \textit{Temporal Grounding},
we employ gpt-oss-120b \citep{openai2025gptoss120bgptoss20bmodel} to generate questions based on any two of the three annotation dimensions including temporal window, artifact type, and facial region,
with the answer derived from the remaining dimension.
The same annotation strategy used in \LI\,
which converts target descriptions into four-option multiple-choice questions,
is applied to produce corresponding QA pairs.
Each question includes a real video option, 
though only the actual authentic video corresponds to that choice. 
Distractors are designed to let models rely on dynamic visual semantics for temporal grounding, 
establishing the necessary basis for robust forgery identification.

\noindent \textbf{Forensic Reasoning.}
For both \textit{Forgery Analysis} and \textit{Final Assessment} tasks, the constructed questions,unlike the previous levels, contain no explicit cue information such as time windows, facial regions, or artifact types.
This requires the model to autonomously establish its reasoning process and effectively perceive forgery patterns throughout the video.
The distractors are carefully designed as partially correct options, 
demanding comprehensive understanding of dynamic visual semantics for accurate discrimination.
Additionally,
we employ the annotation method from \LII\ to construct authentic video samples, ensuring balanced positive and negative examples in the dataset.

\subsection{Human Verification}
All annotations undergo rigorous human verification.
We develop an online validation platform (see Appendix) and recruit volunteers from our institution for the evaluation process.
Each level follows distinct verification protocols.
For \LI, 
validators verify that answers accurately describe the specified facial regions or edge properties visible in the video frames. 
Concerning \LII, they confirm the presence of artifact types and regions mentioned in answers while ensuring distractors reference facial areas absent from the video.
Specifically for temporal grounding questions,
validators must locate the start-end frames corresponding to each answer and verify temporal consistency.
Regarding \LIII, evaluators examine all options against the full video sequence and select the optimal choice,
retaining only those matching the ground-truth annotation. 
Across all levels, questions with unanswerable content or annotation mismatches are discarded,
while those with imperfect distractors or phrasing undergo revision.
This iterative refinement continues until all QA pairs meet our quality standards. 
On average, validation requires 1.5 minutes per question for level 1, 3 minutes for level 2, and 5 minutes for level 3, resulting in a final curated set of 33K QA pairs.

\subsection{FAQ Statistics}
\begin{figure*}[t]
    \centering
    \begin{subfigure}[b]{0.33\linewidth}
        \centering
        \includegraphics[width=\linewidth]{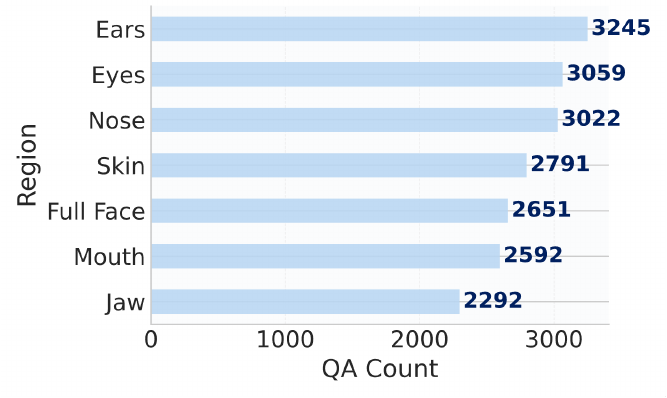}
        \caption{}
        \label{fig:region_distri}
    \end{subfigure}
    \hfill
    \begin{subfigure}[b]{0.33\linewidth}
        \centering
        \includegraphics[width=\linewidth]{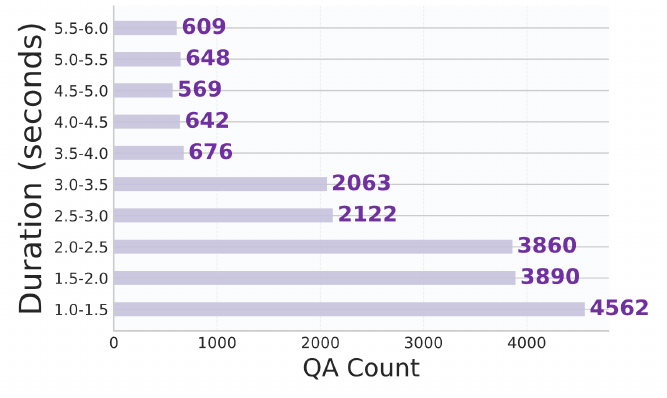}
        \caption{}
        \label{fig:time_distri}
    \end{subfigure}
    \hfill
    \begin{subfigure}[b]{0.33\linewidth}
        \centering
        \includegraphics[width=\linewidth]{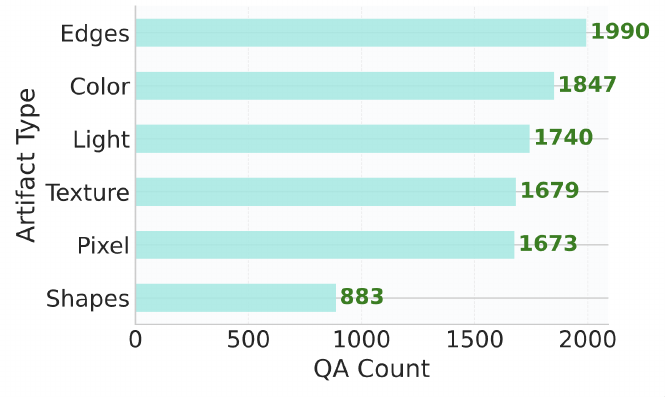}
        \caption{}
        \label{fig:type_distri}
    \end{subfigure}

    \caption{Visualization of data distribution in FAQ across three levels. (a), (b) and (c) shows QA distribution under different facial region, duration and artifact, respectively.}
    \label{fig:faq_data_distribution}
    \vspace{-2mm}
\end{figure*}

We conducted statistical analysis on FAQ to ensure its balance and diversity.
We visualized the data based on the following factors: the length of the video clip; the facial region involved and the artifact type.
See \autoref{fig:faq_data_distribution} for details.

%% file: sec/experiment.tex
\section{Experiment}
\label{sec:experiment}
We evaluate multiple VLMs on our proposed FAQ benchmark.
To support its utility, we also generate the instruction-tuning set (aka, FAQ-IT) using the same pipeline.
Fine-tuning experiments confirm that FAQ-IT leads to effective models, validating our data 
process.
Further ablation studies provide an analysis of the impact of key design choices.

\noindent \textbf{Evaluation models.}
We evaluate a set of VLMs including both open-source and closed-source models. The open-source models comprise InternVL-Chat-V1.2 \citep{InternVL}, LLaVA-1.5 \citep{llava1.5}, DeepSeek-VL \citep{deepseekvl}, LLaVA-InternLM2 \citep{llava_internlm2}, ShareGPT4V \citep{sharegpt4v}, InternVL2 \citep{internvl2}, InternVideo2.5 \citep{internvideo25}, ShareGPT4Video \citep{sharegpt4video}, Qwen3-VL \citep{qwen3vl}, LLaVA-NeXT \citep{llavanext} and Qwen2.5-VL \citep{qwen25vl}. The closed-source models include GPT-4o \citep{gpt4o} and Gemini-2.5-Flash \citep{gemini2.5}.

\noindent \textbf{Training Implementation Details.}
Supervised fine-tuning was conducted on the FAQ-IT dataset with Qwen2.5-VL-7B \citep{qwen25vl} and LLaVA-NeXT-7B \citep{llavanext}.
Specifically, Qwen2.5-VL-7B employs a redesigned Vision Transformer trained from scratch as the vision encoder,
while LLaVA-NeXT-7B utilizes CLIP-ViT-L/336.
We adopt a frozen vision encoder for our training,
based on an empirical finding that unfreezing it yields no significant gains. 
Consequently, we keep the encoder frozen and update the parameters of the visual connector and full-set parameters of the large language model.
We train the models for one epoch using the AdamW optimizer with a weight decay of 0.1,
a global batch size of 16,
and a cosine learning rate scheduler featuring a peak learning rate of 1e-5 and 500 warm-up steps.
All experiments were run on $4\times$ NVIDIA H200 GPU.
See the appendix for more modeling details.

\noindent \textbf{Metrics.}
Unless otherwise noted, we adopt MCQ accuracy as our main evaluation metric following \citep{sida, fakeshield}.
For cross-dataset detection,
it reflects the accuracy of a binary (i.e., ``\textit{real}" or ``\textit{fake}") classification.
For the in-domain FAQ benchmark,
we compute Top-1 accuracy across its three levels,
where a model must exactly match the ground-truth option.
This requires accurately identifying artifacts in forgeries and pristine ``\textit{No artifact}" cases in authentic videos.

\subsection{Results}
\noindent \textbf{Zero-Shot Evaluation.}
We conducted a zero-shot evaluation of 13 VLMs on the proposed FAQ benchmark, as reported in \autoref{tab:performance_on_FAQ}.
The results reveal a significant capability gap in deepfake detection. 
While models show some competency in basic facial perception (level 1),
they struggle with the transition to the spatio-temporal grounding and complex reasoning required at level 2 and level 3.
Notably, the leading commercial models did not achieve top performance,
potentially due to a lack of prior exposure to forensic data in their training.
Among open-source models,
ShareGPT4V-7B delivered the strongest overall results,
a success we attribute to its training data's closer alignment with our benchmark,
which includes landmarks and celebrity images.
In addition, Qwen3-VL-8B achieved the best performance on the spatio-temporal grounding task,
a strength we attribute to its robust video-text alignment capabilities from pre-training.

\begin{table}[t]
\centering
\caption{Average and per-level accuracy (\%) of various VLMs on our proposed FAQ benchmark. Our benchmark evaluates fine-grained reasoning across three levels.}
\label{tab:performance_on_FAQ}
\resizebox{\columnwidth}{!}{
\begin{tabular}{@{}l
                   S[table-format=2.1]
                   S[table-format=2.1]
                   S[table-format=2.1]
                   S[table-format=2.1]@{}}
\toprule
\textbf{Model} & {\textbf{Level 1}} & {\textbf{Level 2}} & {\textbf{Level 3}} & {\textbf{Average}} \\
\midrule
\multicolumn{5}{@{}c}{\textit{Proprietary Large Vision Language Models}} \\
\midrule
GPT-4o            & 26.9 & 27.1 & 13.2 & 22.8 \\
Gemini-2.5-Flash   & 40.0 & 25.6 & 15.3 & 27.8 \\
\midrule
\multicolumn{5}{@{}c}{\textit{Open-sourced Large Vision Language Models}} \\
\midrule
\addlinespace
InternVL-Chat-V1-2   & 29.4 & 21.4 & 13.7 & 21.5 \\
LLaVA-1.5-13b        & 36.6 & 27.2 & 17.3 & 27.3 \\
DeepSeek-VL-7B       & 54.7 & 26.1 & 13.7 & 32.1 \\
LLaVA-InternLM2-7B   & 32.1 & 23.2 & 14.2 & 22.8 \\
ShareGPT4V-7B        & \textbf{73.8} & 20.5 & \textbf{22.3}& \textbf{39.7} \\
InternVL2-8B         & 24.0 & 29.2 & 15.1 & 22.8 \\
InternVideo2.5       & 54.6 & 21.8 & 14.7 & 30.3 \\
ShareGPT4Video-7B    & 50.7 & 19.3 & 21.1 & 30.3 \\
Qwen3-VL-8B          & 45.6 & \textbf{29.4} & 15.0 & 30.3 \\
LLaVA-NeXT-7B        & 40.0 & 29.0 & 21.2 & 30.3 \\
Qwen2.5-VL-7B         & 24.1 & 23.8 & 16.8 & 21.6 \\
\bottomrule
\end{tabular}
}
\vspace{-2mm}
\end{table}

\noindent \textbf{Training with \BenchIT.}
\begin{table}[t] 
\centering
\caption{Ablation study of BenchIT fine-tuning on LLaVA-NeXT and Qwen2.5-VL models across our benchmark levels. VE denotes Vision Encoder. The symbol ${\phi}$ denotes a zero-shot scenario where the model has not been trained. Symbol ${\BenchIT^{\spadesuit}}$ and ${\BenchIT}$ means training models on static-only QA from FAQ and training on full FAQ, respectively.}
\label{tab:train_with_FAQ}
\setlength\tabcolsep{5pt} 
\resizebox{\columnwidth}{!}{%
\begin{tabular}{
    @{}l
    p{17mm}
    l
    l
    l
    l
    @{}
}
\toprule
\textbf{Model} & \textbf{Condition} & \textbf{Level 1} & \textbf{Level 2} & \textbf{Level 3} & \textbf{Average} \\
\midrule
\multirow{3}{*}{Qwen2.5-VL}
& {\small ${\phi}$} & 24.1 & 23.8 & 16.8 & 21.6 \\
& {\small ${\BenchIT^{\spadesuit}}$} & 31.3$_{\uparrow7.2}$ & 21.9$_{\downarrow1.9}$ & 17.9$_{\uparrow1.1}$ & 23.9$_{\uparrow2.3}$ \\
& {\small ${\BenchIT}$} & \textbf{89.9$_{\uparrow65.8}$} & 41.4$_{\uparrow17.6}$ & 25.8$_{\uparrow9.0}$ & 52.4$_{\uparrow30.8}$ \\
\addlinespace
\multirow{3}{*}{LLaVA-NeXT}
& {\small ${\phi}$} & 40.0 & 29.0 & 21.2 & 30.3 \\
& {\small ${\BenchIT^{\spadesuit}}$} & 49.2$_{\uparrow9.2}$ & 28.8$_{\downarrow0.2}$ & 23.3$_{\uparrow2.1}$ & 33.8$_{\uparrow3.5}$ \\
& {\small ${\BenchIT}$} & 88.8$_{\uparrow48.8}$ & \textbf{45.8$_{\uparrow16.8}$} & \textbf{26.5$_{\uparrow5.3}$} & \textbf{53.7$_{\uparrow23.4}$} \\
\bottomrule
\end{tabular}
} 
\vspace{-4mm}
\end{table}
We fine-tuned Qwen2.5-VL and LLaVA-NeXT on the FAQ dataset due to their distinct and widely adopted architectures.
As shown in \autoref{tab:train_with_FAQ}, full-data fine-tuning yielded significant gains across all question tiers.
Notably, LLaVA-NeXT achieved the most substantial improvement (48.8\% average accuracy increase),
compared to 30.79\% for Qwen2.5-VL.
However, training with only static-cue data provided limited and unstable benefits,
particularly for LLaVA-NeXT which saw negligible gains.
We hypothesize that this discrepancy stems from architectural differences,
specifically LLaVA-NeXT's vision encoder and its initially weaker instruction-following capability,
making it more dependent on the diverse spatio-temporal information in our full dataset.

\noindent \textbf{Downstream Deepfake Detection Capabilities.} 
Following conventional metrics,
we evaluate models on FF++ in both zero-shot and fine-tuned settings.
FAQ training leads to significant gains over baselines (details in \autoref{tab:MCQ_indomain} below).
Qwen2.5-VL excels in detection across all manipulation types,
owing to its vision encoder's sensitivity to low-level artifacts.
LLaVA-NeXT's superior video-text alignment and temporal grounding capabilities make it particularly adept at leveraging our FAQ training, which in turn leads to greater gains in overall performance.

\subsection{Ablations}

\noindent \textbf{Cross-Manipulation Evaluations.}
Following the protocol of prior works \cite{altfreezing, FreqBlender, LAA_Net, LSDA}, we evaluate models trained on FAQ on unseen forgery techniques within FF++.
As \autoref{tab:MCQ_indomain} above shows, full training on FAQ–IT yields substantial gains, boosting MCQ accuracy by 20\% to 30\% across most manipulations.
This indicates that FAQ enhances general forgery detection capabilities, rather than overfitting to specific artifacts.
A notable exception is the performance on Face2Face (F2F),
which remains low.
We hypothesize that F2F's forgery artifacts are more temporally subtle and are not adequately captured by our cross-frame sampling strategy, 
limiting the model's ability to learn robust features.

\begin{table}[t]
\centering
\caption{In-domain MCQ accuracy \& In-domain Detection accuracy. Symbol definitions follow \autoref{tab:train_with_FAQ}.}
\label{tab:MCQ_indomain}
\footnotesize
\setlength\tabcolsep{1pt}
\resizebox{\columnwidth}{!}{%
\begin{tabular}{
    @{}l l
    l l l l l l
    @{}
}
\toprule
\textbf{Model} & \textbf{Condition} &
{\textbf{FS}} & {\textbf{NT}} &
{\textbf{F2F}} & {\textbf{DF}} & {\textbf{FSh}} & {\textbf{Avg}} \\
\midrule
\multicolumn{8}{@{}c}{\textit{In-domain MCQ ACC (\%)}} \\
\midrule
\multirow{3}{*}{Qwen2.5-VL}
    & \textbf{${\phi}$}         & 9.7 & 7.4 & 13.5 & 9.6 & 7.7 & 9.6 \\
    & ${\BenchIT^{\spadesuit}}$ & 10.5$_{\uparrow 0.8}$ & 13.3$_{\uparrow 5.8}$ & 12.8$_{\downarrow 0.7}$ & 10.4$_{\uparrow 0.8}$ & 10.7$_{\uparrow 3.0}$ & 11.5$_{\uparrow 1.9}$ \\
    & \textbf{${\BenchIT}$}     & 45.9$_{\uparrow 36.2}$ & 46.7$_{\uparrow 39.3}$ & 24.4$_{\uparrow 10.9}$ & 45.3$_{\uparrow 35.7}$ & 45.3$_{\uparrow 37.6}$ & 41.5$_{\uparrow 31.9}$ \\
\addlinespace
\multirow{3}{*}{LLaVA-NeXT}
    & \textbf{${\phi}$}         & 24.7 & 23.0 & 19.7 & 21.6 & 24.2 & 22.6 \\
    & ${\BenchIT^{\spadesuit}}$ & 27.2$_{\uparrow 2.5}$ & 20.8$_{\downarrow 2.2}$ & 25.0$_{\uparrow 5.3}$ & 22.4$_{\uparrow 0.8}$ & 27.9$_{\uparrow 3.7}$ & 24.7$_{\uparrow 2.1}$ \\
    & \textbf{${\BenchIT}$}     & \textbf{49.3$_{\uparrow 24.6}$} & \textbf{50.6}$_{\uparrow 27.6}$ & \textbf{26.0}$_{\uparrow 6.3}$ & \textbf{49.2}$_{\uparrow 27.6}$ & \textbf{46.9}$_{\uparrow 22.7}$ & 44.4$_{\uparrow 21.8}$ \\
\addlinespace[2pt]
\midrule
\multicolumn{8}{@{}c}{\textit{In-domain Detection ACC (\%)}} \\
\midrule
\multirow{3}{*}{Qwen2.5-VL}
    & \textbf{${\phi}$}         & 17.8 & 11.6 & 20.3 & 36.7  & 27.9 & 22.8 \\
    & ${\BenchIT^{\spadesuit}}$ & 27.2$_{\uparrow 9.4}$ & 34.5$_{\uparrow 22.9}$ & 47.4$_{\uparrow 27.1}$ & 69.2$_{\uparrow 32.5}$ & 49.7$_{\uparrow 21.8}$ & 45.6$_{\uparrow 22.8}$ \\
    & ${\BenchIT}$     & \textbf{71.7}$_{\uparrow 53.9}$ & 59.8$_{\uparrow 48.2}$ & \textbf{74.7}$_{\uparrow 54.4}$ & \textbf{92.7}$_{\uparrow 56.0}$ & 69.9$_{\uparrow 42.0}$ & 73.8$_{\uparrow 51.0}$ \\
\addlinespace
\multirow{3}{*}{LLaVA-NeXT}
    & \textbf{${\phi}$}         & 14.0 & 13.9 & 17.9 & 29.9 & 24.3 & 20.0 \\
    & ${\BenchIT^{\spadesuit}}$ & 29.4$_{\uparrow 15.4}$ & 26.3$_{\uparrow 12.4}$ & 30.7$_{\uparrow 12.8}$ & 42.1$_{\uparrow 12.2}$ & 46.9$_{\uparrow 22.6}$ & 35.1$_{\uparrow 15.1}$ \\
    & \textbf{${\BenchIT}$}     & 64.8$_{\uparrow 50.8}$ & \textbf{60.1}$_{\uparrow 46.2}$ & 69.7$_{\uparrow 51.8}$ & 78.2$_{\uparrow 48.3}$ & \textbf{73.7}$_{\uparrow 49.4}$ & 69.3$_{\uparrow 49.3}$ \\
\bottomrule
\end{tabular}
}
\vspace{-4mm}
\end{table}

\noindent \textbf{Robustness Analysis.}
The robustness of the SFT-trained models was evaluated against baseline methods across a range of video compression levels, including original quality, c23, and c40.
Here, c23 and c40 denote the Constant Rate Factor (CRF) values used in H.264 video encoding, representing light and heavy compression levels, respectively.
Results in \autoref{tab:MCQ_indomain} show that both models maintain high accuracy at original quality and under light compression (c23),
but suffer severe degradation under high compression (c40).
LLaVA-NeXT remains the top performer across most settings.
We hypothesize that aggressive compression destroys subtle spatial forgery artifacts, thereby depriving the visual encoders of the critical cues necessary for reliable detection.

\begin{table}[t]
\centering
\caption{In-domain MCQ accuracy under different compressions on FF++. Symbol definitions follow \autoref{tab:train_with_FAQ}.} 
\label{tab:compression}
\footnotesize
\setlength\tabcolsep{5pt}
\resizebox{\columnwidth}{!}{%
\begin{tabular}{
    @{}l l
    S[table-format=2.1]
    S[table-format=2.1]
    S[table-format=2.1]
    S[table-format=2.1]
    S[table-format=2.1]@{}
}
\toprule
\textbf{Model} & \textbf{Com.} &
{\textbf{FS}} & {\textbf{NT}} &
{\textbf{F2F}} & {\textbf{DF}} & {\textbf{FSh}} \\
\midrule

\multirow{2}{*}{Qwen2.5-VL$_{\BenchIT}$}
    & c23       & 45.9 & 46.7 & 24.4 & 45.3 & 45.3 \\
    & c40       & 41.8 & 42.2 & 20.9 & 40.1 & 40.4 \\
    & original  & 46.1 & 46.5 & \textbf{27.2} & 47.3 & 46.0 \\
\addlinespace
\multirow{2}{*}{LLaVA-NeXT$_{\BenchIT}$}
    & c23       & \textbf{49.3} & 50.6 & 26.0 & 49.2 & \textbf{46.9} \\
    & c40       & 44.1 & 46.2 & 21.2 & 46.1 & 42.9 \\
    & original  & 49.1 & \textbf{51.9} & 26.9 & \textbf{50.9} & 43.7 \\
\bottomrule
\end{tabular}   
} 
\end{table}

\noindent \textbf{Analysis on Forgery Properties.}
As presented in \autoref{tab:different_artifact_types}, evaluation across different artifact types indicates robust overall performance from our trained models. 
Note that unlike \autoref{tab:train_with_FAQ}, this evaluation utilizes only the ``FAKE" video samples and not all of them.
A notable exception is \textit{Texture Inconsistencies},
where models fine-tuned on our non-full FAQ data (i.e., \BenchIT$^{\spadesuit}$) show reduced accuracy. 
We hypothesize that detecting such fine-grained,
low-level artifacts demands a higher degree of visual acuity and texture discrimination,
which may exceed the current capabilities of general-purpose VLMs.
Furthermore, empirical results indicate that Qwen2.5-VL attains the best performance after full FAQ fine-tuning,
outperforming LLaVA-NeXT.
This result underscores its superior instruction-following capability among current open-source VLMs.

\begin{table}[t]
\centering
\caption{MCQ performance for different artifact types in deepfake videos. Symbol definitions follow \autoref{tab:train_with_FAQ}.}
\label{tab:different_artifact_types}
\footnotesize
\setlength\tabcolsep{2pt}
\resizebox{\columnwidth}{!}{%
\begin{tabular}{
    @{}l
    p{17mm}
    l l l l l l
    @{}
}
\toprule
\textbf{Model} & \textbf{Condition} &
\textbf{Edges} & \textbf{Color} & \textbf{Light} &
\textbf{Texture} & \textbf{Pixel} & \textbf{Shapes} \\
\midrule

\multirow{3}{*}{Qwen2.5-VL}
& {\small ${\phi}$} 
    & 11.3 & 8.4 & 4.5 & 8.0 & 13.2 & 9.0 \\
& {\small ${\BenchIT^{\spadesuit}}$} 
    & 13.5 & 9.1 & 7.9 & 10.9 & 14.3 & 12.1 \\
& {\small ${\BenchIT}$} 
    & \textbf{52.9} & \textbf{45.9} & 46.2 & \textbf{43.1} & \textbf{50.2} & 44.3 \\

\addlinespace

\multirow{3}{*}{LLaVA-NeXT}
& {\small ${\phi}$} 
    & 9.8 & 11.1 & 14.9 & 16.6 & 10.7 & 11.4 \\
& {\small ${\BenchIT^{\spadesuit}}$} 
    & 12.8 & 13.9 & 18.8 & 16.4 & 14.3 & 13.4 \\
& {\small ${\BenchIT}$} 
    & 45.7 & 42.1 & \textbf{47.9} & 42.1 & 43.9 & \textbf{46.8} \\

\bottomrule
\end{tabular}
}
\vspace{-4mm}
\end{table}

\noindent \textbf{Cross-Dataset Evaluations.}
To assess the generalization ability of our FAQ-trained models, we conduct open-ended inference on three unseen deepfake datasets including Celeb-DF (CDF \citep{cdf}), DeeperForensics (DFo \citep{dfo}), and WildDeepfake (WDF \citep{wild_deepfake}). 
Following the CLIP's prompt engineering \cite{CLIP}, 
we use the question template ``\textit{Does this video contain forged information?}" to guide the model to output a ``\textit{yes}" or ``\textit{no}" answer.
This allows the model to make determinations about any customized unseen video.
The results are reported in \autoref{tab:cross_dataset_ablation}.
After training on FAQ, both Qwen2.5-VL and LLaVA-NeXT have a noticeable improvement on detection accuracy especially on CDF.
We speculate that the model performs relatively better on CDF because the forgery types are more standardized and the synthesis process is more controlled. 
In contrast, DFo involves greater variability and WDF is collected from the web, resulting in greater data diversity. 
The model struggles to adapt to unknown forgeries.
\vspace{1mm}
\begin{table}[t]
\centering
\caption{Cross-dataset generalization accuracy (\%) and F1-score (\%) on CDF \citep{cdf}, DFo \citep{dfo}, and WDF \citep{wild_deepfake} with different training settings. Symbol definitions follow \autoref{tab:train_with_FAQ}.}
\label{tab:cross_dataset_ablation}
\footnotesize
\setlength\tabcolsep{5pt}
\resizebox{\columnwidth}{!}{%
\begin{tabular}{
    @{}l l
    llllll
    @{}
}
\toprule
\multicolumn{2}{c}{\textbf{Training Setup}} &
\multicolumn{2}{c}{\textbf{CDF}} &
\multicolumn{2}{c}{\textbf{DFo}} &
\multicolumn{2}{c}{\textbf{WDF}} \\
\cmidrule(lr){3-4}
\cmidrule(lr){5-6}
\cmidrule(lr){7-8}
\textbf{Model} & \textbf{Training Data} &
{\textbf{Acc}} & {\textbf{F1}} &
{\textbf{Acc}} & {\textbf{F1}} &
{\textbf{Acc}} & {\textbf{F1}} \\
\midrule
\multirow{3}{*}{Qwen2.5-VL}
    & \textbf{${\phi}$}         & 17.9 & 13.8 & 14.5 & 11.1 & 10.8 & 8.9 \\
    & ${\BenchIT^{\spadesuit}}$ & 49.1 & 59.1 & 54.7 & 57.9 & 40.9 & 54.8 
    \\
    & \textbf{${\BenchIT}$}     & \textbf{73.3} & \textbf{71.0} & \textbf{78.6} & 70.1 & \textbf{65.9} & 59.3 \\
\addlinespace
\multirow{3}{*}{LLaVA-NeXT}
    & \textbf{${\phi}$}         & 12.3 & 10.2 & 10.5 & 9.5 & 8.7 & 7.6 \\
    & \textbf{${\BenchIT}^{\spadesuit}$}     & 47.2 & 49.0 & 52.3 & 54.8 & 39.8 & 49.6 \\
    & ${\BenchIT}$ & 72.9 & 69.1 & 77.8 & \textbf{72.3} & 63.1 & \textbf{62.9} \\
\bottomrule
\end{tabular}%
}
\vspace{-2mm}
\end{table}

\begin{figure}[t]
    \centering 
    \includegraphics[width=0.47\textwidth]{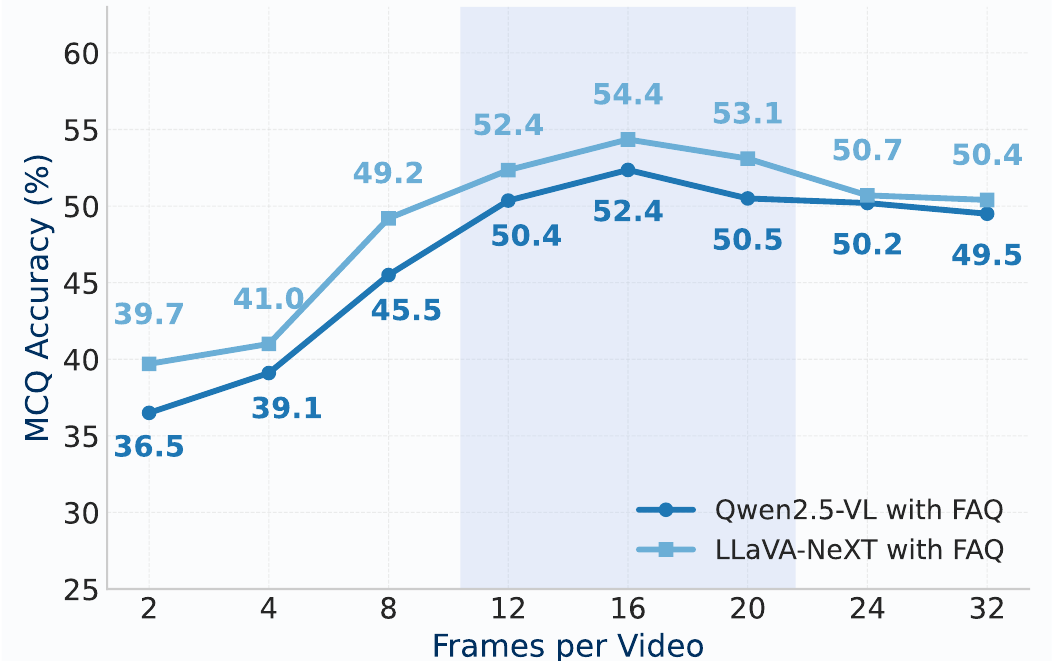} 
    \caption{Effect of frame sampling strategies.}
    \label{fig:frame_ablation} 
\end{figure}

\noindent \textbf{Effect of Frame Sampling Strategies.}
We ablate the video frame sampling strategy to assess its impact on forgery detection performance. Models are trained and evaluated with different frame counts per video: $\{2, 4, 8, 12, 16, 24\}$.
As illustrated in \autoref{fig:frame_ablation},
model accuracy peaks at 16 frames.
We hypothesize that with fewer than 16 frames,
the model lacks sufficient temporal context for robust feature learning. 
Beyond this point, performance degrades, likely due to redundant visual information from uniform sampling, which can distract the model and dilute spatiotemporal cues.

\noindent \textbf{Supervision Design and Shortcut Analysis.}
To further validate FAQ, we conduct ablations on LLaVA-NeXT along robustness, training strategy, and open-ended evaluation (Table R1). 
Under perturbations, performance is stable to brightness shifts (-1.6\%) but degrades under compression (-4.6\%) and more severely under Gaussian blur. 
Since blur suppresses high-frequency inter-frame inconsistencies, this sensitivity confirms reliance on fine-grained temporal artifacts rather than static cues.
For training strategy, multi-stage and Lvl-3-only supervision yield moderate gains, but remain far below the proposed Mixed SFT. 
The gap suggests staged training induces partial catastrophic forgetting, whereas mixed supervision preserves multimodal alignment while enforcing temporal aggregation. 
Thus, performance gains arise from stable curriculum design rather than supervision scale alone.
To exclude MCQ bias, we evaluate open-ended QA. 
GPT-4o achieves 13.2\%, the baseline VLM 18.5\%, and our model 49.1\%, closely matching its MCQ accuracy. 
The consistency rules out option elimination heuristics and confirms genuine temporally grounded reasoning.

\begin{table}[t]
  \centering
  \scriptsize
  \renewcommand{\arraystretch}{1} 
  \renewcommand{\thetable}{R1}
  \setlength{\tabcolsep}{2pt} 
  
  \setlength{\aboverulesep}{0pt} 
  \setlength{\belowrulesep}{0pt}
  
  \caption{\textbf{New Exp on LLaVA-NeXT.} \textbf{Left:} Robustness. \textbf{Mid:} Training Strategy. \textbf{Right:} Open-Ended QA.}
  \label{tab:new_exp}
  
  \begin{tabular}{lc @{\hspace{4mm}} lc @{\hspace{4mm}} lc}
    \toprule
    \multicolumn{2}{c}{\textbf{Exp 1: Robustness}} & \multicolumn{2}{c}{\textbf{Exp 2: Strategy}} & \multicolumn{2}{c}{\textbf{Exp 3: Open-Ended QA}} \\
    \cmidrule(r){1-2} \cmidrule(lr){3-4} \cmidrule(l){5-6}
    Cond. & Acc & Strategy & Acc & Model & Score \\ 
    \midrule
    Original & 54.4\% & Baseline & 30.3\% & Random & $\approx$0.0\% \\
    Bright. & $52.8\%_{\bad{-1.6}}$ & 
    Multi-Stage & $35.6\%_{\text{\good{+5.3}}}$ &
    GPT-4o & 13.2\% \\
    Compr. & $49.8\%_{\bad{-4.6}}$ &
    Lvl 3 Only & $37.8\%_{\text{\good{+7.5}}}$ & 
    Baseline & 18.5\% \\
    Blur & $46.2\%_{\textbf{\bad{-8.2}}}$ & \textbf{Mixed} & $\mathbf{54.4\%_{\text{\good{+24.1}}}}$ & \textbf{Ours} & $\mathbf{49.1\%}$ \\
    \bottomrule
  \end{tabular}
\end{table}

%% file: sec/conclusion.tex
\vspace{-2mm}
\section{Conclusion}
In this paper, we present FAQ, a QA benchmark focused on leveraging the potential of temporal inconsistencies in deepfake videos for VLMs to learn.
Starting with manual annotation, we use a comprehensive process to convert static annotations into questions and answers related to temporal information. 
Thanks to carefully designed task definitions and distractors, these QAs effectively evaluate models' temporal reasoning capability for deepfake videos.
We validated that our dataset can effectively enhance model's performance on deepfake detection tasks and temporal reasoning tasks through training.
\label{sec:conclusion}

%% file: sec/X_suppl.tex
\clearpage
\setcounter{page}{1}
\maketitlesupplementary
\lstdefinestyle{promptstyle}{
  basicstyle=\ttfamily\small,
  breaklines=true,
  frame=single,
  backgroundcolor=\color{gray!5},
  keywordstyle={},
  showstringspaces=false,
}

\section{Data Filtering}
\begin{figure}[t]
    \centering 
    \includegraphics[width=\columnwidth]{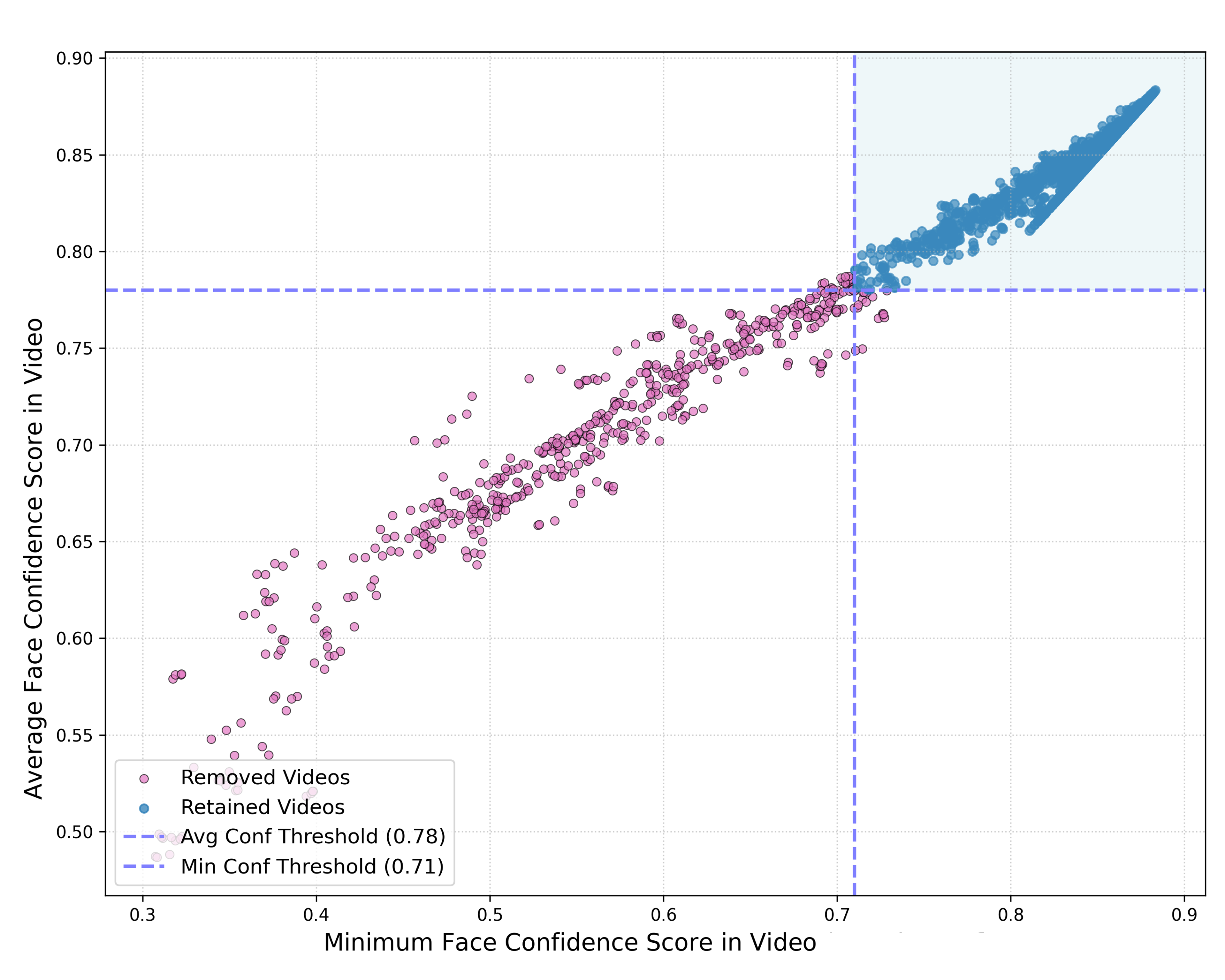} 
    \caption{Distribution of Videos across Face Detection Confidence Metrics and the Filtering Boundary. Each point represents a video from the original set. The filtering boundary (dashed light blue lines) separates the retained high-quality samples (blue points) from the removed low-quality samples (light purple points), demonstrating the rigorous exclusion of videos with unstable or weak face detections.}
    \label{fig:filtering} 
    \vspace{-0.2cm}
\end{figure}

We performed frame-level face detection on all video files in the FF++ dataset, and utilized the confidence scores from face detection as the benchmark for data cleaning.
A video was deemed satisfactory and thus retained only if its average face confidence score exceeded 0.78 and its minimum per-frame confidence score was above 0.71. This two-pronged criterion ensures not only a high overall quality but, crucially, prevents the inclusion of videos with transient but severe detection failures (e.g., due to heavy occlusion or poor camera angle). As visualized in \autoref{fig:filtering}, this process effectively removes approximately 10\% of the initial unsatisfactory videos, yielding a refined set of over 4500 high-quality samples.
\begin{table}[t]
\centering
\caption{Ablation study on data filtering ratios. The table reports \textbf{(Top)} Overall benchmark accuracy including real cases; and \textbf{(Bottom)} In-domain and Cross-domain tasks. \textbf{90\%} corresponds to our proposed setting.}
\label{tab:single_column_ablation_v2}
\footnotesize
\setlength\tabcolsep{2pt}
\resizebox{\columnwidth}{!}{%
\begin{tabular}{
    @{}l l
    ccccc
    @{}
}
\toprule
\multicolumn{7}{@{}c}{\textbf{\textit{Overall Benchmark Accuracy (\%)}}} \\
\midrule
\textbf{Model} & \textbf{Data Ratio} &
\textbf{Level 1} & \textbf{Level 2} & \textbf{Level 3} & \multicolumn{2}{c}{\textbf{Average}} \\
\midrule
\multirow{6}{*}{Qwen2.5-VL}
    & 100\% (Raw) & \cellcolor{gray!15}\textbf{90.3} & 40.1 & 23.5 & \multicolumn{2}{c}{51.3} \\
    & 90\% (Ours) & 89.9 & \cellcolor{gray!15}\textbf{41.4} & \cellcolor{gray!15}\textbf{25.8} & \multicolumn{2}{c}{\cellcolor{gray!15}\textbf{52.4}} \\
    & 80\% & 89.7 & 41.0 & 24.9 & \multicolumn{2}{c}{51.9} \\
    & 70\% & 88.1 & 39.5 & 23.2 & \multicolumn{2}{c}{50.3} \\
    & 60\% & 85.4 & 36.8 & 20.1 & \multicolumn{2}{c}{47.4} \\
    & 50\% & 81.2 & 33.2 & 18.5 & \multicolumn{2}{c}{44.3} \\
\addlinespace
\multirow{6}{*}{LLaVA-NeXT}
    & 100\% (Raw) & \cellcolor{gray!15}\textbf{89.2} & 43.9 & 24.2 & \multicolumn{2}{c}{52.4} \\
    & 90\% (Ours) & 88.8 & 45.8 & \cellcolor{gray!15}\textbf{26.5} & \multicolumn{2}{c}{\cellcolor{gray!15}\textbf{53.7}} \\
    & 80\% & 88.5 & \cellcolor{gray!15}\textbf{45.9} & 26.1 & \multicolumn{2}{c}{53.5} \\
    & 70\% & 86.9 & 43.5 & 24.4 & \multicolumn{2}{c}{51.6} \\
    & 60\% & 83.5 & 40.1 & 21.8 & \multicolumn{2}{c}{48.5} \\
    & 50\% & 79.4 & 36.5 & 19.6 & \multicolumn{2}{c}{45.2} \\
\addlinespace[4pt]
\toprule
\multicolumn{7}{@{}c}{\textbf{\textit{In-domain (Avg) \& Cross-domain Generalization (Acc)}}} \\
\midrule
\textbf{Model} & \textbf{Data Ratio} &
\textbf{In-MCQ} & \textbf{In-Det} & \textbf{CDF} & \textbf{DFo} & \textbf{WDF} \\
\midrule
\multirow{6}{*}{Qwen2.5-VL}
    & 100\% (Raw) & 40.2 & 71.5 & 71.5 & 76.2 & 63.8 \\
    & 90\% (Ours) & \cellcolor{gray!15}\textbf{41.5} & 73.8 & \cellcolor{gray!15}\textbf{73.3} & \cellcolor{gray!15}\textbf{78.6} & \cellcolor{gray!15}\textbf{65.9} \\
    & 80\% & 41.7 & \cellcolor{gray!15}\textbf{74.2} & 72.8 & 78.1 & 64.9 \\
    & 70\% & 39.8 & 72.1 & 69.5 & 76.4 & 62.5 \\
    & 60\% & 36.5 & 67.2 & 66.1 & 72.8 & 58.9 \\
    & 50\% & 33.2 & 60.8 & 62.4 & 68.5 & 55.1 \\
\addlinespace
\multirow{6}{*}{LLaVA-NeXT}
    & 100\% (Raw) & 43.1 & 67.5 & 70.8 & 75.9 & 61.2 \\
    & 90\% (Ours) & \cellcolor{gray!15}\textbf{44.4} & \cellcolor{gray!15}\textbf{69.3} & 72.9 & \cellcolor{gray!15}\textbf{77.8} & \cellcolor{gray!15}\textbf{63.1} \\
    & 80\% & 44.2 & 69.0 & \cellcolor{gray!15}\textbf{73.1} & 77.2 & 62.5 \\
    & 70\% & 42.5 & 66.8 & 69.1 & 74.8 & 60.4 \\
    & 60\% & 39.1 & 62.4 & 65.3 & 70.5 & 57.2 \\
    & 50\% & 35.8 & 58.1 & 61.7 & 66.2 & 53.8 \\
\bottomrule
\end{tabular}
}
\vspace{-2mm}
\end{table}

To empirically validate the efficacy of our data curation strategy, we conducted a sensitivity analysis by varying the retention ratio (100\%, 90\%, 80\%, and 70\%) based on the confidence scores of the filtering mechanism. The results are shown in \autoref{tab:single_column_ablation_v2}.

Comparing the 90\% setting against the full dataset (100\%), we observe a measurable degradation in the latter, particularly in in-domain detection accuracy (e.g., a decline from 73.8\% to 71.5\% for Qwen2.5-VL). This suggests that the bottom 10\% of samples introduce detrimental noise rather than informative features. These low-confidence samples likely contain severe visual degradation or alignment errors, which impair the optimization process by confusing the model's decision boundaries.

Conversely, increasing the filtering strictness to 80\% and 70\% negatively impacts cross-domain generalization (CDF, DFo, WDF). This decline indicates that while the bottom 10\% constitutes noise, the subsequent data segments contribute significantly to the distributional diversity of the training set. Excluding these samples reduces the feature variance necessary for the model to learn robust representations against unseen manipulation types. Consequently, the 90\% threshold establishes an optimal equilibrium between maximizing data quality and preserving the semantic diversity required for effective generalization.

\section{Prompt Details}
To facilitate reproducibility and provide comprehensive implementation details, we present the exact prompts employed to guide the LLM in our pipeline. 

\noindent \textbf{Prompt for Description Parsing.}
This prompt guides the LLM to decompose raw video descriptions into atomic, structured annotations.
\begin{lstlisting}[style=promptstyle,caption={Prompt for Description Parsing}]
You are a data processing assistant.
Your task is to extract fine-grained "atomic annotations" from a raw video description.

Input Description:
"{raw_description}" (e.g., "The video shows a person whose nose looks blurry and the mouth area has inconsistent skin textures.")

Instructions:
1. Decompose the description into separate atomic units.
2. Each unit must describe exactly ONE facial region and ONE specific artifact type.
3. Map the facial region to one of these categories: [Eyes, Nose, Mouth, Jaw, Ears].
4. Map the artifact type to standard forensic terms: [Blur, Color Distortion, Texture Inconsistency, Boundary Artifacts, Lighting Anomaly].
5. Ignore subjective adjectives (e.g., "weird", "ugly") and focus on visual evidence.

Output Format (JSON):
[
  {"region": "Nose", "artifact": "Blur"},
  {"region": "Mouth", "artifact": "Texture Inconsistency"}
]
\end{lstlisting}

\noindent \textbf{Prompt for Generating QA in Level 1.}
This prompt constructs Region Perception and Edge Perception tasks with visually plausible distractors.
\vspace{-0.1cm}
\begin{lstlisting}[style=promptstyle, caption={Prompt for Generating QA in Level 1}]
You are a Visual QA generator. Generate a binary choice question (A/B) to test the perception of artifacts in a specific facial region.

Input Data:
- Region: "{region}" (e.g., Skin)
- State: "{state}" (e.g., Pixelated / Real)

Instructions:
1. Question: Must start with "Observe the {region}." followed by "Which option best describes its visual quality?"
2. Options:
   - Option A: Describe a high-quality, natural appearance (e.g., "Clear, well-defined...").
   - Option B: Describe the artifact or distortion (e.g., "Pixelated, blurry, or fake-looking").
   - Randomize the order of A and B in the final output.

Example Output:
{
  "question": "Observe the Skin. Which option best describes its visual quality?",
  "options": {
    "A": "Clear, well-defined, with natural appearance.",
    "B": "Blurred, distorted, or fake-looking."
  },
  "answer": "B"
}
\end{lstlisting}

\noindent \textbf{Prompt for Generating QA in Level 2.}
This prompt generates Type Understanding, Region Grounding, and Temporal Grounding tasks by masking one dimension of the annotation.
\begin{lstlisting}[style=promptstyle, caption={Prompt for Generating QA in Level 2}]
You are a generic QA generator. Given a video timeline and artifact details, generate a temporal grounding question.

Input Data:
- Artifact: "{artifact}"
- Manipulation Time: "{start}s-{end}s" (or "None" if Real)
- Task: "Identify the time range of the manipulation."

Instructions:
1. Question: "During which time range does the face appear most realistic, without manipulation artifacts?" (or inverse for fake).
2. Options (Generate 5 options A-E):
   - Include the Correct Time Range.
   - Include 3 Distractor Time Ranges (non-overlapping or partial).
   - ALWAYS include one "Authentic" option: "The video appears harmonious and consistent, showing no signs of manipulation. It is authentic."

Example Output:
{
  "question": "During which time range does the face appear most realistic, without manipulation artifacts?",
  "options": {
    "A": "8.40s-10.72s",
    "B": "10.72s-13.44s", 
    "C": "3.80s-7.04s",
    "D": "The video appears harmonious and consistent, showing no signs of manipulation. It is authentic.",
    "E": "10.44s-12.88s"
  },
  "answer": "D"
}
\end{lstlisting}
\noindent \textbf{Prompt for Generating QA in Level 3.}
This prompt constructs complex options combining temporal, spatial, and artifact attributes.
\begin{lstlisting}[style=promptstyle,caption={Prompt for Generating QA in Level 3}]
You are a Forensic Expert. Generate a multiple-choice question where the options are detailed forensic descriptions.

Input Trace:
- Correct: Time="{t_start}-{t_end}", Region="{region}", Artifact="{type}"

Instructions:
1. Question: "Which of the following descriptions best matches the manipulated content of this video segment?"
2. Option Format: MUST follow this template: "Between {t_start} and {t_end}, the {region} shows signs of {type}."
3. Distractors:
   - Generate options that mix up the dimensions (e.g., Correct Time but Wrong Region, or Correct Region but Wrong Type).
   - Include an "Authentic" option if applicable.

Example Output:
{
  "question": "<video>Which of the following descriptions best matches the manipulated content of this video segment?",
  "options": {
    "A": "Between 4.44s and 4.96s, the Mouth shows signs of Blurred edges.",
    "B": "The visual sequence is continuous and believable... It looks real.",
    "C": "Between 5.29s and 5.87s, the Ears shows signs of Pixelation.",
    "D": "Between 4.44s and 4.96s, the Ears shows signs of Pixelation." 
  },
  "answer": "D"
}
\end{lstlisting}

\section{Modeling Details}
The models and corresponding parameters used for training and evaluation are presented in \autoref{tab:training_hyperparameters}.

\begin{table}[htbp]
\centering
\caption{Hyperparameters for Training and Inference}
\label{tab:training_hyperparameters}
\resizebox{\columnwidth}{!}{%
\begin{tabular}{lc}
\toprule
\textbf{Configuration} & \textbf{Value} \\
\midrule
\multicolumn{2}{l}{\textit{Video Input Preprocessing}} \\
\midrule
Max Pixels per Video & $50176$ \\
Max Frames per Video & $36$ \\
Input Video Frame Resolution & $448 \times 448$ \\
\midrule
\multicolumn{2}{l}{\textit{Training / Fine-tuning Configuration}} \\
\midrule
Total Training Epochs & $1$ \\
Optimizer & \texttt{AdamW} ($\beta_1=0.9, \beta_2=0.95$) \\
Weight Decay & $0.1$ \\
Global Batch Size & $16$ \\
Batch Size (per GPU) & $4$ \\
Total GPUs Used & $4$ \\
Learning Rate (LR) & $1 \times 10^{-5}$ \\
LR Scheduler & Cosine Decay \\
Warm-up Steps & $500$ \\
\bottomrule
\end{tabular}%
}
\end{table}

\section{Annotation Platform}
\label{sec:annotation}
\begin{figure*}[t]
    \centering 
    \includegraphics[width=\textwidth]{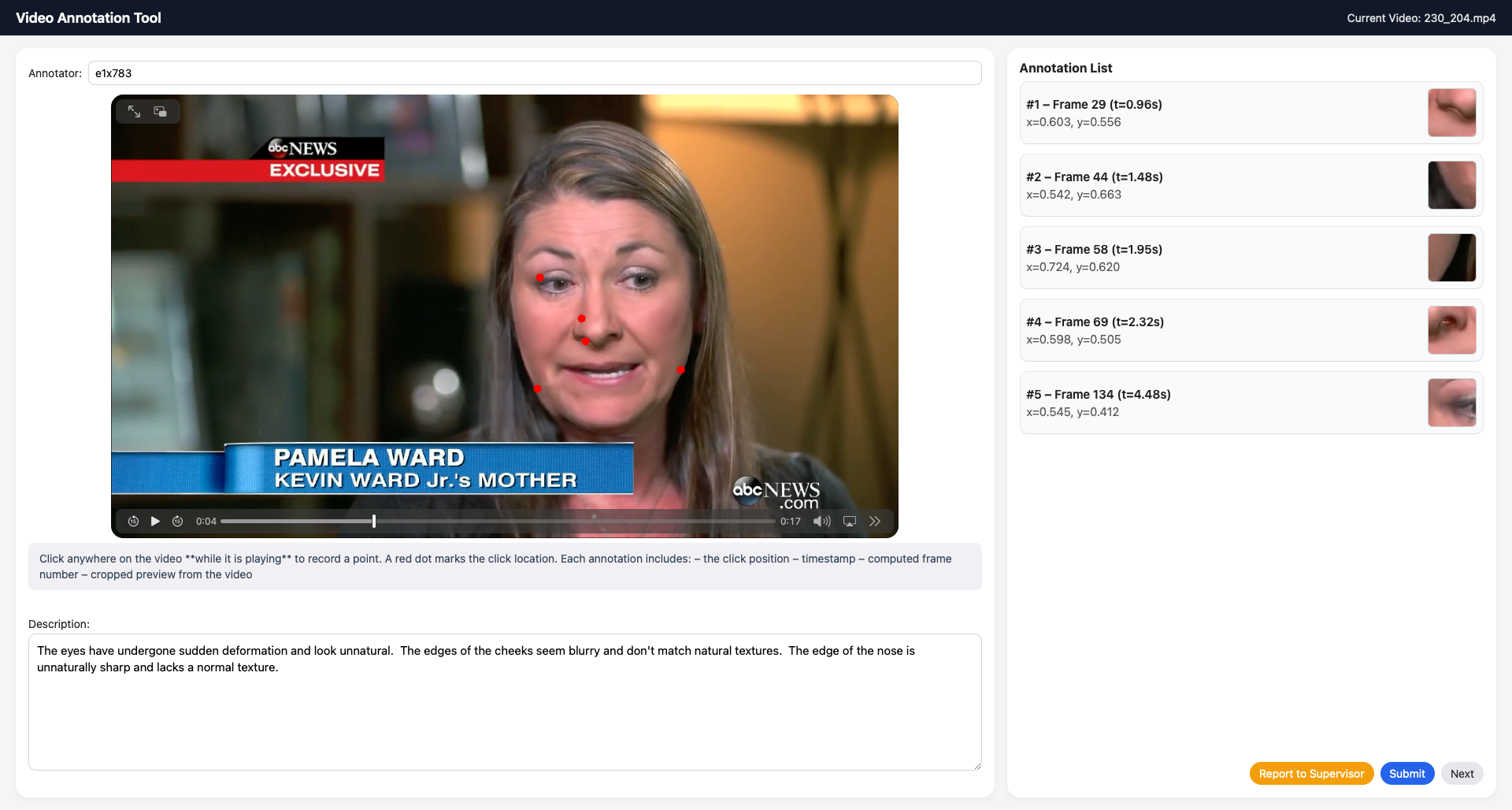} 
    \caption{Overview of our custom-developed video annotation interface. The platform is designed to capture fine-grained spatiotemporal forgery traces. The central player allows annotators to mark dynamic artifacts (indicated by red dots) in real-time, recording precise $(x, y, t)$ coordinates. The right panel displays the sequential list of annotated clicks with corresponding frame previews, while the bottom text field captures a detailed natural language description of the observed manipulation anomalies.}
    \label{fig:annotation} 
\end{figure*}

We have developed an online human annotation platform and recruited volunteers within the organization, all holding at least a bachelor's degree, to perform the annotations. The annotation platform is displayed as shown in \autoref{fig:annotation}.

\section{Verification Platform}
\begin{figure*}[t]
    \centering 
    \includegraphics[width=\textwidth]{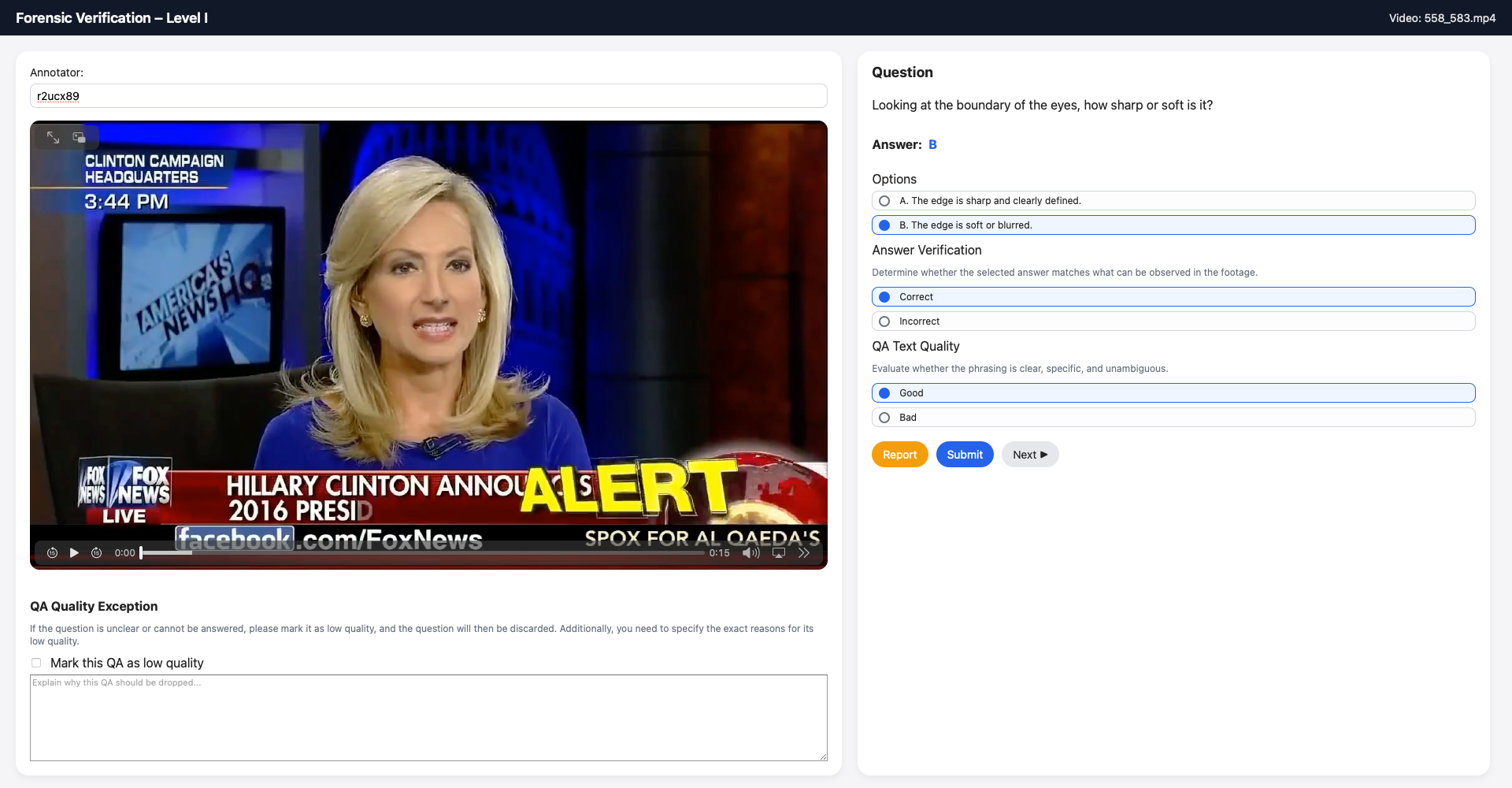} 
    \caption{User interface of our verification platform for Level 1 (Facial Perception). To ensure high data quality, validators execute a dual-check protocol: (1) Answer Verification, ensuring the selected option (e.g., ``soft or blurred edges'') accurately reflects the visual artifacts observed in the video; and (2) Text Quality Assessment, confirming that the phrasing is unambiguous. The ``Exception'' panel at the bottom allows validators to flag and discard samples with severe hallucinations or low-quality generation.}
    \label{fig:verification_l1} 
\end{figure*}

\begin{figure*}[t]
    \centering 
    \includegraphics[width=\textwidth]{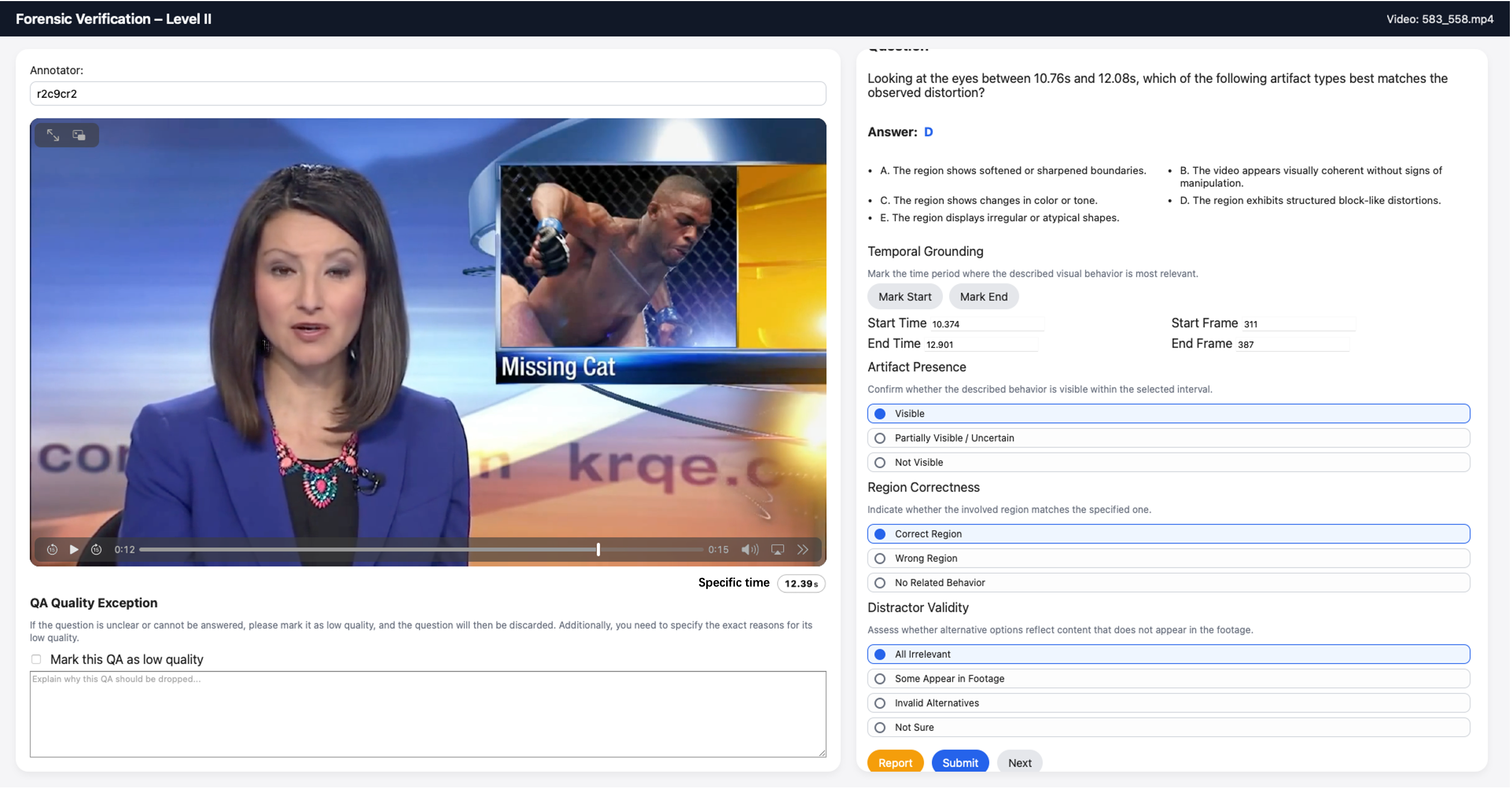} 
    \caption{The verification interface for Level 2 (Temporal Deepfake Grounding). In contrast to Level 1, this stage requires rigorous temporal verification. Validators must manually use the ``Mark Start'' and ``Mark End'' buttons to define the precise boundaries of the artifact, ensuring the timestamps in the generated QA match the actual video content. Additional checks for Artifact Presence, Region Correctness, and Distractor Validity are implemented to prevent ambiguous or incorrectly grounded samples.}
    \label{fig:verification_l2} 
\end{figure*}

\begin{figure*}[t]
    \centering 
    \includegraphics[width=\textwidth]{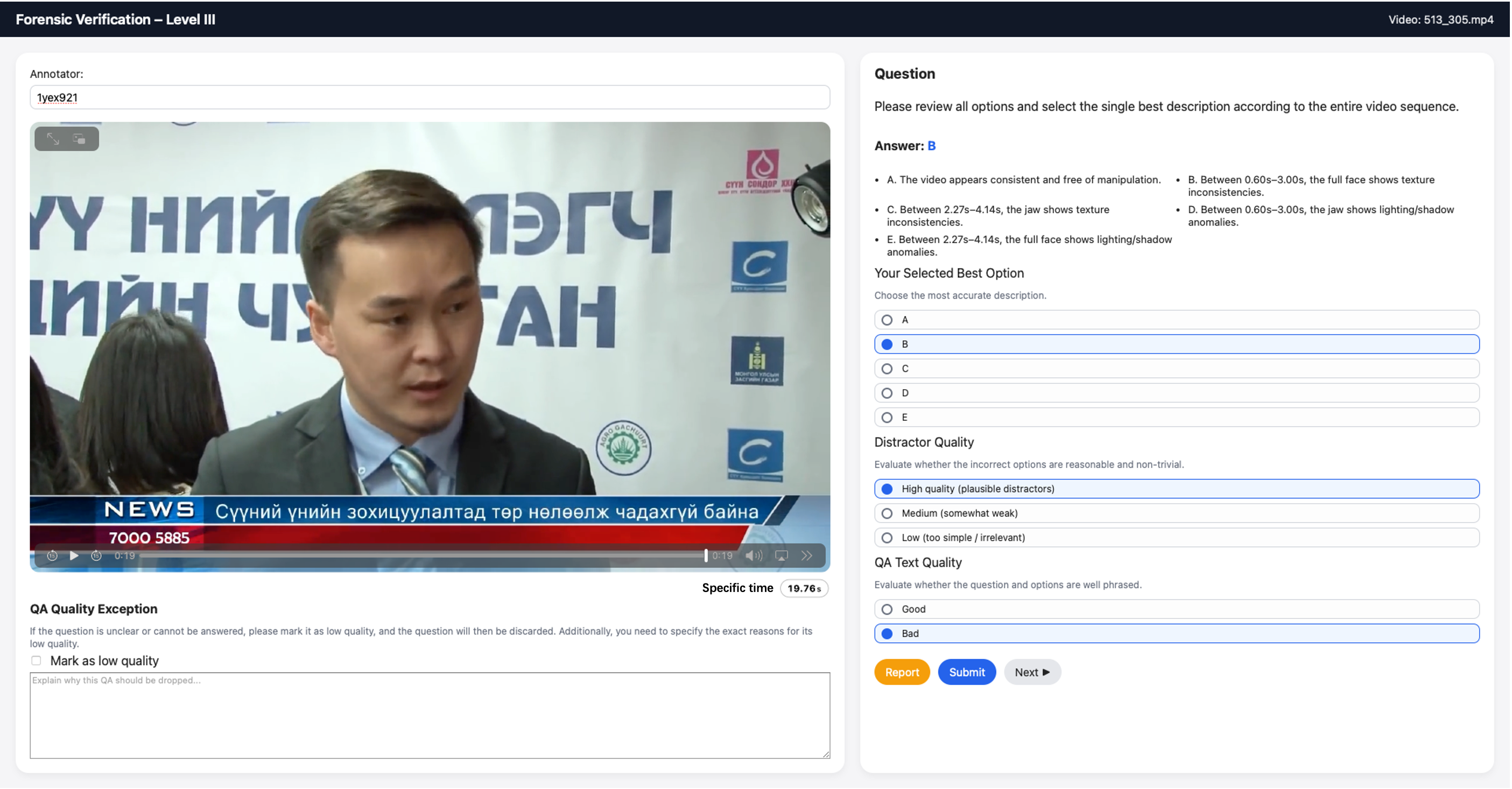} 
    \caption{The verification interface for Level 3 (Forensic Reasoning). Unlike lower-level tasks, this stage requires validators to perform a holistic assessment of the entire video sequence. Validators are tasked with: (1) independently selecting the most accurate description among complex options to confirm the ground truth's validity; and (2) evaluating Distractor Quality, ensuring that incorrect options are ``plausible and non-trivial'' to guarantee the task effectively challenges the model's reasoning capabilities.}
    \label{fig:verification_l3} 
\end{figure*}

Following the annotation process \autoref{sec:annotation}, we have established an online data validation platform to assess the usability of the QA data we have created. 
The verification platform is displayed as shown in \autoref{fig:verification_l1}, \autoref{fig:verification_l2} and \autoref{fig:verification_l3}.
For each level, we have designed distinct verification processes.

%% file: main.bib
@String(CVPR= {IEEE Conf. Comput. Vis. Pattern Recog.})

@String(ICCV= {Int. Conf. Comput. Vis.})

@String(CVPR  = {CVPR})

@String(ICCV  = {ICCV})

@inproceedings{ddvqa,
  title={Common sense reasoning for deepfake detection},
  author={Zhang, Yue and Colman, Ben and Guo, Xiao and Shahriyari, Ali and Bharaj, Gaurav},
  booktitle={European conference on computer vision},
  pages={399--415},
  year={2024},
  organization={Springer}
}

@inproceedings{vlffd,
  title={Towards General Visual-Linguistic Face Forgery Detection},
  author={Ke Sun and Shen Chen and Taiping Yao and Haozhe Yang and Xiaoshuai Sun and Shouhong Ding and Rongrong Ji},
  booktitle={CVPR},
  year={2025}
}

@inproceedings{CLIP,
  title = 	 {Learning Transferable Visual Models From Natural Language Supervision},
  author =       {Radford, Alec and Kim, Jong Wook and Hallacy, Chris and Ramesh, Aditya and Goh, Gabriel and Agarwal, Sandhini and Sastry, Girish and Askell, Amanda and Mishkin, Pamela and Clark, Jack and Krueger, Gretchen and Sutskever, Ilya},
  booktitle = 	 {Proceedings of the 38th International Conference on Machine Learning},
  pages = 	 {8748--8763},
  year = 	 {2021},
  editor = 	 {Meila, Marina and Zhang, Tong},
  volume = 	 {139},
  series = 	 {Proceedings of Machine Learning Research},
  month = 	 {18--24 Jul},
  publisher =    {PMLR},
  url = 	 {https://proceedings.mlr.press/v139/radford21a.html}
}

@misc{gpt4v,
      title={The Dawn of LMMs: Preliminary Explorations with GPT-4V(ision)}, 
      author={Zhengyuan Yang and Linjie Li and Kevin Lin and Jianfeng Wang and Chung-Ching Lin and Zicheng Liu and Lijuan Wang},
      year={2023},
      eprint={2309.17421},
      archivePrefix={arXiv},
      primaryClass={cs.CV},
      url={https://arxiv.org/abs/2309.17421}, 
}

@inproceedings{InternVL,
  author={Chen, Zhe and Wu, Jiannan and Wang, Wenhai and Su, Weijie and Chen, Guo and Xing, Sen and Zhong, Muyan and Zhang, Qinglong and Zhu, Xizhou and Lu, Lewei and Li, Bin and Luo, Ping and Lu, Tong and Qiao, Yu and Dai, Jifeng},
  booktitle={2024 IEEE/CVF Conference on Computer Vision and Pattern Recognition (CVPR)}, 
  title={Intern VL: Scaling up Vision Foundation Models and Aligning for Generic Visual-Linguistic Tasks}, 
  year={2024},
  volume={},
  number={},
  pages={24185-24198},
  doi={10.1109/CVPR52733.2024.02283}}

@misc{qwen25vl,
      title={Qwen2.5-VL Technical Report}, 
      author={Shuai Bai and Keqin Chen and Xuejing Liu and Jialin Wang and Wenbin Ge and Sibo Song and Kai Dang and Peng Wang and Shijie Wang and Jun Tang and Humen Zhong and Yuanzhi Zhu and Mingkun Yang and Zhaohai Li and Jianqiang Wan and Pengfei Wang and Wei Ding and Zheren Fu and Yiheng Xu and Jiabo Ye and Xi Zhang and Tianbao Xie and Zesen Cheng and Hang Zhang and Zhibo Yang and Haiyang Xu and Junyang Lin},
      year={2025},
      eprint={2502.13923},
      archivePrefix={arXiv},
      primaryClass={cs.CV},
      url={https://arxiv.org/abs/2502.13923}, 
}

@inproceedings{LSDA,
    author    = {Yan, Zhiyuan and Luo, Yuhao and Lyu, Siwei and Liu, Qingshan and Wu, Baoyuan},
    title     = {Transcending Forgery Specificity with Latent Space Augmentation for Generalizable Deepfake Detection},
    booktitle = {Proceedings of the IEEE/CVF Conference on Computer Vision and Pattern Recognition (CVPR)},
    month     = {June},
    year      = {2024},
    pages     = {8984-8994}
}

@inproceedings{LAA_Net,
  author={NGUYEN, Dat and MEJRI, Nesryne and SINGH, Inder Pal and KULESHOVA, Polina and ASTRID, Marcella and KACEM, Anis and GHORBEL, Enjie and AOUADA, Djamila},
  booktitle={2024 IEEE/CVF Conference on Computer Vision and Pattern Recognition (CVPR)}, 
  title={LAA-Net: Localized Artifact Attention Network for Quality-Agnostic and Generalizable Deepfake Detection}, 
  year={2024},
  volume={},
  number={},
  pages={17395-17405},
  doi={10.1109/CVPR52733.2024.01647}
}

@inproceedings{M2F2_det,
    author    = {Guo, Xiao and Song, Xiufeng and Zhang, Yue and Liu, Xiaohong and Liu, Xiaoming},
    title     = {Rethinking Vision-Language Model in Face Forensics: Multi-Modal Interpretable Forged Face Detector},
    booktitle = {Proceedings of the IEEE/CVF Conference on Computer Vision and Pattern Recognition (CVPR)},
    month     = {June},
    year      = {2025},
    pages     = {105-116}
}

@inproceedings{FreqBlender,
 author = {Li, Hanzhe and Zhou, Jiaran and Li, Yuezun and Wu, Baoyuan and Li, Bin and Dong, Junyu},
 booktitle = {Advances in Neural Information Processing Systems},
 editor = {A. Globerson and L. Mackey and D. Belgrave and A. Fan and U. Paquet and J. Tomczak and C. Zhang},
 pages = {44965--44988},
 publisher = {Curran Associates, Inc.},
 title = {FreqBlender: Enhancing DeepFake Detection by Blending Frequency Knowledge},
 url = {https://proceedings.neurips.cc/paper_files/paper/2024/file/4f92d2f498b88f1bd43732312272967a-Paper-Conference.pdf},
 volume = {37},
 year = {2024}
}

@misc{FF++,
      title={FaceForensics++: Learning to Detect Manipulated Facial Images}, 
      author={Andreas Rössler and Davide Cozzolino and Luisa Verdoliva and Christian Riess and Justus Thies and Matthias Nießner},
      year={2019},
      eprint={1901.08971},
      archivePrefix={arXiv},
      primaryClass={cs.CV},
      url={https://arxiv.org/abs/1901.08971}, 
}

@article{altfreezing,
  title={AltFreezing for More General Video Face Forgery Detection},
  author={Zhendong Wang and Jianmin Bao and Wen-gang Zhou and Weilun Wang and Houqiang Li},
  journal={2023 IEEE/CVF Conference on Computer Vision and Pattern Recognition (CVPR)},
  year={2023},
  pages={4129-4138},
  url={https://api.semanticscholar.org/CorpusID:259229566}
}

@article{Deepfake_in_politics,
 ISSN = {00157120, 23277793},
 URL = {https://www.jstor.org/stable/26798018},
 author = {Robert Chesney and Danielle Citron},
 journal = {Foreign Affairs},
 number = {1},
 pages = {pp. 147--155},
 publisher = {Council on Foreign Relations},
 title = {Deepfakes and the New Disinformation War: The Coming Age of Post-Truth Geopolitics},
 urldate = {2025-10-28},
 volume = {98},
 year = {2019}
}

@inbook{Deepfake_in_finance,
author = {Khan, Rizwan and Taqi, Mohd and Afzal, Mohd},
year = {2024},
month = {07},
pages = {91-120},
title = {Deepfakes in Finance: Unraveling the Threat Landscape and Detection Challenges},
isbn = {9798369352984},
doi = {10.4018/979-8-3693-5298-4.ch006}
}

@misc{wild_deepfake,
      title={WildDeepfake: A Challenging Real-World Dataset for Deepfake Detection}, 
      author={Bojia Zi and Minghao Chang and Jingjing Chen and Xingjun Ma and Yu-Gang Jiang},
      year={2024},
      eprint={2101.01456},
      archivePrefix={arXiv},
      primaryClass={cs.CV},
      url={https://arxiv.org/abs/2101.01456}, 
}

@inproceedings{fakeshield,
title={FakeShield: Explainable Image Forgery Detection and Localization via Multi-modal Large Language Models},
author={Xu, Zhipei and Zhang, Xuanyu and Li, Runyi and Tang, Zecheng and Huang, Qing and Zhang, Jian},
booktitle={International Conference on Learning Representations},
year={2025}
}

@inproceedings{sida,
  author       = {Zhenglin Huang and Jinwei Hu and Xiangtai Li and Yiwei He and Xingyu Zhao and Bei Peng and Baoyuan Wu and Xiaowei Huang and Guangliang Cheng},
  title        = {{SIDA:} Social Media Image Deepfake Detection, Localization and Explanation
                  with Large Multimodal Model},
  booktitle    = {{IEEE/CVF} Conference on Computer Vision and Pattern Recognition(CVPR) 2025},
  year         = {2025},
}

@inproceedings{sft,
 author = {Ouyang, Long and Wu, Jeffrey and Jiang, Xu and Almeida, Diogo and Wainwright, Carroll and Mishkin, Pamela and Zhang, Chong and Agarwal, Sandhini and Slama, Katarina and Ray, Alex and Schulman, John and Hilton, Jacob and Kelton, Fraser and Miller, Luke and Simens, Maddie and Askell, Amanda and Welinder, Peter and Christiano, Paul F and Leike, Jan and Lowe, Ryan},
 booktitle = {Advances in Neural Information Processing Systems},
 editor = {S. Koyejo and S. Mohamed and A. Agarwal and D. Belgrave and K. Cho and A. Oh},
 pages = {27730--27744},
 publisher = {Curran Associates, Inc.},
 title = {Training language models to follow instructions with human feedback},
 url = {https://proceedings.neurips.cc/paper_files/paper/2022/file/b1efde53be364a73914f58805a001731-Paper-Conference.pdf},
 volume = {35},
 year = {2022}
}

@article{sharegpt4v,
  title={ShareGPT4V: Improving Large Multi-Modal Models with Better Captions},
  author={Chen, Lin and Li, Jisong and Dong, Xiaoyi and Zhang, Pan and He, Conghui and Wang, Jiaqi and Zhao, Feng and Lin, Dahua},
  journal={arXiv preprint arXiv:2311.12793},
  year={2023}
}

@article{sharegpt4video,
  title={ShareGPT4Video: Improving Video Understanding and Generation with Better Captions},
  author={Chen, Lin and Wei, Xilin and Li, Jinsong and Dong, Xiaoyi and Zhang, Pan and Zang, Yuhang and Chen, Zehui and Duan, Haodong and Lin, Bin and Tang, Zhenyu and others},
  journal={arXiv preprint arXiv:2406.04325},
  year={2024}
}

@article{internvl2,
    title={How Far Are We to GPT-4V? Closing the Gap to Commercial Multimodal Models with Open-Source Suites},
    author={Chen, Zhe and Wang, Weiyun and Tian, Hao and Ye, Shenglong and Gao, Zhangwei and Cui, Erfei and Tong, Wenwen and Hu, Kongzhi and Luo, Jiapeng and Ma, Zheng and others},
    journal={arXiv preprint arXiv:2404.16821},
    year={2024}
}

@misc{internvideo25,
      title={InternVideo2.5: Empowering Video MLLMs with Long and Rich Context Modeling}, 
      author={Yi Wang and Xinhao Li and Ziang Yan and Yinan He and Jiashuo Yu and Xiangyu Zeng and Chenting Wang and Changlian Ma and Haian Huang and Jianfei Gao and Min Dou and Kai Chen and Wenhai Wang and Yu Qiao and Yali Wang and Limin Wang},
      year={2025},
      eprint={2501.12386},
      archivePrefix={arXiv},
      primaryClass={cs.CV},
      url={https://arxiv.org/abs/2501.12386}, 
}

@misc{deepseekvl,
      title={DeepSeek-VL: Towards Real-World Vision-Language Understanding}, 
      author={Haoyu Lu and Wen Liu and Bo Zhang and Bingxuan Wang and Kai Dong and Bo Liu and Jingxiang Sun and Tongzheng Ren and Zhuoshu Li and Hao Yang and Yaofeng Sun and Chengqi Deng and Hanwei Xu and Zhenda Xie and Chong Ruan},
      year={2024},
      eprint={2403.05525},
      archivePrefix={arXiv},
      primaryClass={cs.AI},
      url={https://arxiv.org/abs/2403.05525}, 
}

@misc{qwen3vl,
  author = {QwenLM},
  title = {{Qwen3-VL}},
  year = {2025},
  publisher = {GitHub},
  howpublished = {\url{https://github.com/QwenLM/Qwen3-VL}},
  note = {Accessed: 2025-11-07}
}

@misc{llavanext,
    title={LLaVA-NeXT: Improved reasoning, OCR, and world knowledge},
    url={https://llava-vl.github.io/blog/2024-01-30-llava-next/},
    author={Liu, Haotian and Li, Chunyuan and Li, Yuheng and Li, Bo and Zhang, Yuanhan and Shen, Sheng and Lee, Yong Jae},
    month={January},
    year={2024}
}

@misc{llava1.5,
      title={Improved Baselines with Visual Instruction Tuning}, 
      author={Haotian Liu and Chunyuan Li and Yuheng Li and Yong Jae Lee},
      year={2024},
      eprint={2310.03744},
      archivePrefix={arXiv},
      primaryClass={cs.CV},
      url={https://arxiv.org/abs/2310.03744}, 
}

@misc{gpt4o,
      title={GPT-4o System Card}, 
      author={OpenAI team},
      year={2024},
      eprint={2410.21276},
      archivePrefix={arXiv},
      primaryClass={cs.CL},
      url={https://arxiv.org/abs/2410.21276}, 
}

@misc{gemini2.5,
      title={Gemini 2.5: Pushing the Frontier with Advanced Reasoning, Multimodality, Long Context, and Next Generation Agentic Capabilities}, 
      author={Gemini team},
      year={2025},
      eprint={2507.06261},
      archivePrefix={arXiv},
      primaryClass={cs.CL},
      url={https://arxiv.org/abs/2507.06261}, 
}

@misc{dfo,
      title={DeeperForensics-1.0: A Large-Scale Dataset for Real-World Face Forgery Detection}, 
      author={Liming Jiang and Ren Li and Wayne Wu and Chen Qian and Chen Change Loy},
      year={2020},
      eprint={2001.03024},
      archivePrefix={arXiv},
      primaryClass={cs.CV},
      url={https://arxiv.org/abs/2001.03024}, 
}

@misc{openai2025gptoss120bgptoss20bmodel,
      title={gpt-oss-120b \& gpt-oss-20b Model Card}, 
      author={OpenAI},
      year={2025},
      eprint={2508.10925},
      archivePrefix={arXiv},
      primaryClass={cs.CL},
      url={https://arxiv.org/abs/2508.10925}, 
}

@misc{cdf,
      title={Celeb-DF: A Large-scale Challenging Dataset for DeepFake Forensics}, 
      author={Yuezun Li and Xin Yang and Pu Sun and Honggang Qi and Siwei Lyu},
      year={2020},
      eprint={1909.12962},
      archivePrefix={arXiv},
      primaryClass={cs.CR},
      url={https://arxiv.org/abs/1909.12962}, 
}

@inproceedings{yolov8,
  author={Varghese, Rejin and M., Sambath},
  booktitle={2024 International Conference on Advances in Data Engineering and Intelligent Computing Systems (ADICS)}, 
  title={YOLOv8: A Novel Object Detection Algorithm with Enhanced Performance and Robustness}, 
  year={2024},
  volume={},
  number={},
  pages={1-6},
  doi={10.1109/ADICS58448.2024.10533619}}

@inproceedings{Forensics-Bench,
  author={Wang, Jin and Lv, Chenghui and Li, Xian and Dong, Shichao and Li, Huadong and Yao, Kelu and Li, Chao and Shao, Wenqi and Luo, Ping},
  booktitle={2025 IEEE/CVF Conference on Computer Vision and Pattern Recognition (CVPR)}, 
  title={Forensics-Bench: A Comprehensive Forgery Detection Benchmark Suite for Large Vision Language Models}, 
  year={2025},
  volume={},
  number={},
  pages={4233-4245},
  doi={10.1109/CVPR52734.2025.00400}}

@article{dlib, author = {King, Davis E.}, title = {Dlib-ml: A Machine Learning Toolkit}, year = {2009}, issue_date = {12/1/2009}, publisher = {JMLR.org}, volume = {10}, issn = {1532-4435}, journal = {J. Mach. Learn. Res.}, month = dec, pages = {1755–1758}, numpages = {4} }

@misc{genimage,
      title={GenImage: A Million-Scale Benchmark for Detecting AI-Generated Image}, 
      author={Mingjian Zhu and Hanting Chen and Qiangyu Yan and Xudong Huang and Guanyu Lin and Wei Li and Zhijun Tu and Hailin Hu and Jie Hu and Yunhe Wang},
      year={2023},
      eprint={2306.08571},
      archivePrefix={arXiv},
      primaryClass={cs.CV}
}

@article{genvideo,
      title={DeMamba: AI-Generated Video Detection on Million-Scale GenVideo Benchmark},
      author={Haoxing Chen and Yan Hong and Zizheng Huang and Zhuoer Xu and Zhangxuan Gu and Yaohui Li and Jun Lan and Huijia Zhu and Jianfu Zhang and Weiqiang Wang and Huaxiong Li},
      journal={arXiv preprint arXiv:2405.19707},
      year={2024}
}

@inproceedings{nltk,
    title = "{NLTK}: The Natural Language Toolkit",
    author = "Bird, Steven  and
      Loper, Edward",
    booktitle = "Proceedings of the {ACL} Interactive Poster and Demonstration Sessions",
    month = jul,
    year = "2004",
    address = "Barcelona, Spain",
    publisher = "Association for Computational Linguistics",
    url = "https://aclanthology.org/P04-3031/",
    pages = "214--217"
}

@misc{llava_internlm2,
    title={XTuner: A Toolkit for Efficiently Fine-tuning LLM},
    author={XTuner Contributors},
    howpublished = {\url{https://github.com/InternLM/xtuner}},
    year={2023}
}

@inproceedings{deepshield,
    author    = {Cai, Yinqi and Li, Jichang and Li, Zhaolun and Chen, Weikai and Lan, Rushi and Xie, Xi and Luo, Xiaonan and Li, Guanbin},
    title     = {DeepShield: Fortifying Deepfake Video Detection with Local and Global Forgery Analysis},
    booktitle = {Proceedings of the IEEE/CVF International Conference on Computer Vision (ICCV)},
    month     = {October},
    year      = {2025},
    pages     = {12524-12534}
}

@inproceedings{fakeradar,
    author    = {Li, Zhaolun and Li, Jichang and Cai, Yinqi and Chen, Junye and Luo, Xiaonan and Li, Guanbin and Lan, Rushi},
    title     = {FakeRadar: Probing Forgery Outliers to Detect Unknown Deepfake Videos},
    booktitle = {Proceedings of the IEEE/CVF International Conference on Computer Vision (ICCV)},
    month     = {October},
    year      = {2025},
    pages     = {13382-13392}
}

@misc{fakesformer,
      title={Vulnerability-Aware Spatio-Temporal Learning for Generalizable Deepfake Video Detection}, 
      author={Dat Nguyen and Marcella Astrid and Anis Kacem and Enjie Ghorbel and Djamila Aouada},
      year={2025},
      eprint={2501.01184},
      archivePrefix={arXiv},
      primaryClass={cs.CV},
      url={https://arxiv.org/abs/2501.01184}, 
}

@misc{dfgdcg,
      title={Towards More General Video-based Deepfake Detection through Facial Component Guided Adaptation for Foundation Model}, 
      author={Yue-Hua Han and Tai-Ming Huang and Kai-Lung Hua and Jun-Cheng Chen},
      year={2025},
      eprint={2404.05583},
      archivePrefix={arXiv},
      primaryClass={cs.CV},
      url={https://arxiv.org/abs/2404.05583}, 
}

@misc{gan,
      title={Progressive Growing of GANs for Improved Quality, Stability, and Variation}, 
      author={Tero Karras and Timo Aila and Samuli Laine and Jaakko Lehtinen},
      year={2018},
      eprint={1710.10196},
      archivePrefix={arXiv},
      primaryClass={cs.NE},
      url={https://arxiv.org/abs/1710.10196}, 
}

@misc{rombach2022highresolutionimagesynthesislatent,
      title={High-Resolution Image Synthesis with Latent Diffusion Models}, 
      author={Robin Rombach and Andreas Blattmann and Dominik Lorenz and Patrick Esser and Björn Ommer},
      year={2022},
      eprint={2112.10752},
      archivePrefix={arXiv},
      primaryClass={cs.CV},
      url={https://arxiv.org/abs/2112.10752}, 
}

@misc{kang2023scalingganstexttoimagesynthesis,
      title={Scaling up GANs for Text-to-Image Synthesis}, 
      author={Minguk Kang and Jun-Yan Zhu and Richard Zhang and Jaesik Park and Eli Shechtman and Sylvain Paris and Taesung Park},
      year={2023},
      eprint={2303.05511},
      archivePrefix={arXiv},
      primaryClass={cs.CV},
      url={https://arxiv.org/abs/2303.05511}, 
}

@misc{styleflow,
      title={Exploiting Style Latent Flows for Generalizing Deepfake Video Detection}, 
      author={Jongwook Choi and Taehoon Kim and Yonghyun Jeong and Seungryul Baek and Jongwon Choi},
      year={2024},
      eprint={2403.06592},
      archivePrefix={arXiv},
      primaryClass={cs.CV},
      url={https://arxiv.org/abs/2403.06592}, 
}

@misc{scaleupgan,
      title={Scaling up GANs for Text-to-Image Synthesis}, 
      author={Minguk Kang and Jun-Yan Zhu and Richard Zhang and Jaesik Park and Eli Shechtman and Sylvain Paris and Taesung Park},
      year={2023},
      eprint={2303.05511},
      archivePrefix={arXiv},
      primaryClass={cs.CV},
      url={https://arxiv.org/abs/2303.05511}, 
}

@misc{marra2018gansleaveartificialfingerprints,
      title={Do GANs leave artificial fingerprints?}, 
      author={Francesco Marra and Diego Gragnaniello and Luisa Verdoliva and Giovanni Poggi},
      year={2018},
      eprint={1812.11842},
      archivePrefix={arXiv},
      primaryClass={cs.CV},
      url={https://arxiv.org/abs/1812.11842}, 
}
